# Context-Adaptive Inference: A Unified Statistical and Foundation-Model View

## Authors

**Yue Yao**
0009-0000-8195-3943 · YueYao-stat
Department of Statistics, University of Wisconsin-Madison

**Jingyun Jia**
0009-0006-3241-3485 · Clouddelta
Department of Statistics, University of Wisconsin-Madison

**Dong Liu**
0009-0009-6815-8297 · NoakLiu
Department of Computer Science, Yale University

**Jiaqi Wang**
0009-0003-8531-9490 · w-jiaqi
Paul G. Allen School of Computer Science & Engineering, University of Washington

**Yixin Yang**
0009-0008-1436-5584 · Jessego5
Department of Computer Sciences, University of Wisconsin-Madison

**Eric P. Xing**
0000-0002-3683-4280 · ericxing
Machine Learning Department, Carnegie Mellon University; Mohamed bin Zayed University of Artificial Intelligence

**Caleb N. Ellington** ✉
0000-0001-7029-8023 · cnellington · probablybots
Computational Biology Department, Carnegie Mellon University

**Baiheng Chen**
0000-0001-8554-3984 · BaihengChen
Department of Statistics, University of Wisconsin-Madison

**Rikhil Rao**
0009-0003-5221-1125 · rikhilr
Department of Computer Science, University of Wisconsin - Madison

**Samuel Wales-McGrath**
0009-0008-5405-2646 · Samuel-WM
Department of Computer Science and Engineering, The Ohio State University

**Zhiyuan Li**
0009-0006-6016-7381 · LZYEIL
Department of Computer Sciences, University of Wisconsin-Madison

**Ben Lengerich** ✉
0000-0001-8690-9554 · blengerich · ben_lengerich
Department of Statistics, University of Wisconsin-Madison

✉ — Correspondence possible via GitHub Issues on github.com/AdaptInfer/context-review or email to Caleb N. Ellington <cellingt@cs.cmu.edu> and Ben Lengerich <lengerich@wisc.edu>.

# Contents

# Abstract

Modern predictive systems are expected to adapt their behavior to the specific situation they are facing. A clinical model should not treat every patient the same; a retrieval-augmented model should change its answer when given different evidence; a mixture-of-experts model should route different inputs to different experts. We call this capability **context-adaptive inference**: before predicting, the system uses information about the current context to specialize its parameters or computation for that instance.

This article provides a unified view of context-adaptive inference across three traditions that are usually treated separately: (i) explicit adaptation in statistics (e.g. varying-coefficient models, local regression, hierarchical sharing), (ii) rapid task-specific adaptation in meta-learning and transfer, and (iii) implicit adaptation in large foundation models via prompting, retrieval, and expert routing. We formalize these approaches under a common objective: to map context $c$ to adapted parameters $\theta(c)$, then to predict via $f(x; \theta(c))$. Under squared loss, linear prediction heads, and fixed features, we prove that explicit parameter adaptation and implicit routing are mathematically equivalent to kernel ridge regression on joint features of inputs and context. Building on this bridge, we propose practical design principles and evaluation metrics including adaptation-efficiency, routing stability, and context-specific robustness to guide when to specialize, how to constrain that specialization, and how to audit context-adaptive models in deployment. Finally, we identify open problems in identifiability, robustness under distribution shift, and efficient large-scale adaptation, outlining design principles for methods that are scalable, reliable, and transparent in real-world settings.

# 1. Introduction

A convenient simplifying assumption in statistical modeling is that observations are independent and identically distributed (i.i.d.). This assumption allows us to use a single model to make predictions across all data points. But in practice, this assumption rarely holds. Data are collected across different individuals, environments, and tasks—each with their own characteristics, constraints, and dynamics. When the i.i.d. assumption breaks down, using a single global model can obscure meaningful heterogeneity.

Even when a global model fits aggregate data well, it can hide systematic failure modes. When one subpopulation dominates, the fit is biased toward it and the minority is underrepresented in both fit and prediction (mode collapse). Small but extreme groups can distort a parameter-averaged fit, as in ordinary least squares (outlier sensitivity). When several subpopulations are equally represented, the global model may fit none of them, converging instead to a non-existent average case (phantom populations). Most consequentially, aggregate trends can reverse within subgroups, so the global fit misrepresents the true relationship entirely (hidden confounding, or Simpson's paradox; Figure 1).

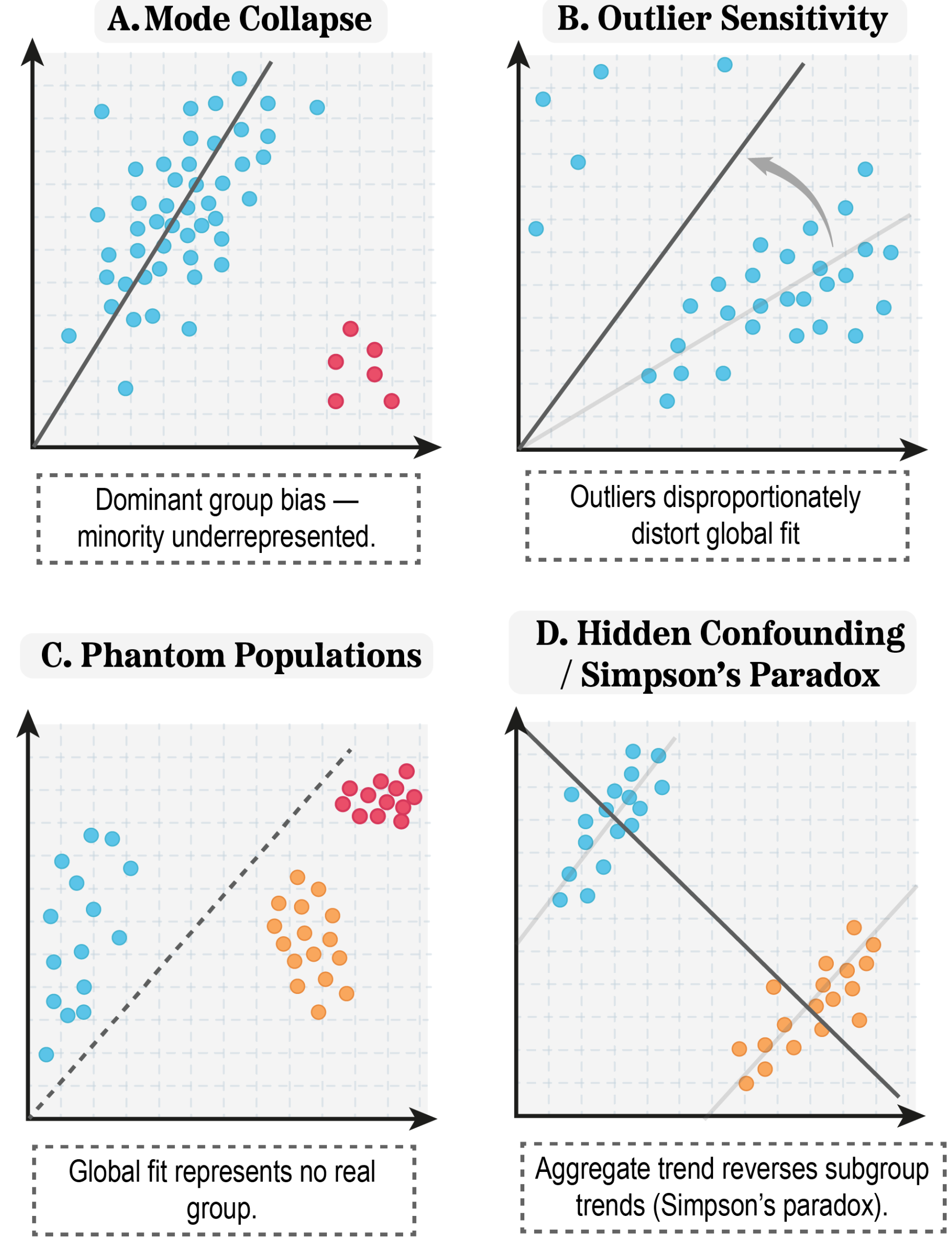

**Figure 1:** Failure Modes of Population Models. Illustrative schematics of common failure types when fitting a single global model to heterogeneous data. (A) **Mode Collapse**: the dominant group drives the fit, underrepresenting the minority. (B) **Outlier Sensitivity**: extreme points distort the global line, shifting predictions away from the majority. (C) **Phantom Populations**: the global fit represents no actual subgroup, but an artificial average. (D) **Hidden Confounding / Simpson's Paradox**: aggregate trends reverse subgroup trends, obscuring true relationships.

The last case is the classical demonstration that heterogeneity cannot be resolved from the outcome data alone: recovering the true subgroup relationships requires conditioning on the contextual side information that distinguishes these groups, i.e. the "no unmeasured confounding" assumption. These are just a few examples based on statistical primitives that illustrate how heterogeneity in data confounds traditional estimators and undermines basic machine learning assumptions.

To model under heterogeneity, a growing class of methods aim to make inference adaptive to context. These include varying-coefficient models in statistics, transfer and meta-learning in machine learning, and in-context learning in large foundation models. Though these approaches arise from different traditions, they share a common goal: to use contextual information—whether covariates, environments, or support sets—to inform sample-specific inference.

Throughout this work, we rely on statistical primitives to characterize heterogeneity in data, build intuition about what types of models can learn under heterogeneity, and understand how context-adaptive learning occurs. These insights reveal overarching principles in model design and data composition that enable context-adaptive inference, which apply simultaneously to statistical and large-scale foundation models. Figure 2 summarizes how statistics, meta-learning, and foundation models are all unified by a context→parameters view, later formalized in Eq. 1.

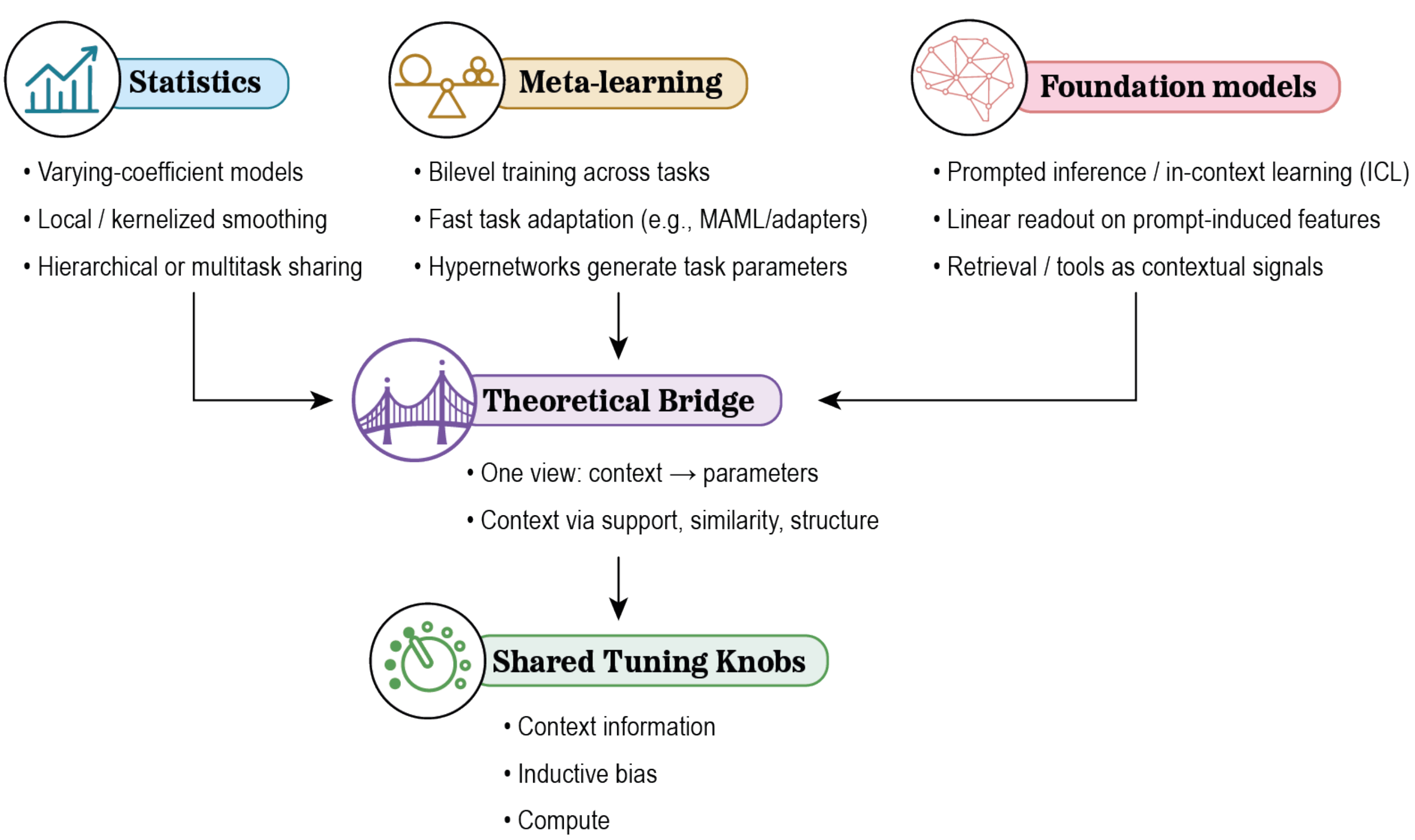

**Figure 2:** Overview of the theoretical bridge. Three traditions—Statistics (varying-coefficients, local smoothing, hierarchical sharing), Meta-learning (bilevel training, fast adaptation, hypernetworks), and Foundation models (prompted inference / in-context learning)—feed into a unified context→parameters view. The bridge formalizes this connection and highlights shared tuning knobs: context information, inductive bias, and compute.

Heterogeneity, the basic idea that samples are not identically distributed, is the foundation for the problems and methods addressed in this work. We formalize heterogeneity by assuming each observation $x_i$ is drawn from a sample-specific distribution governed by parameters $\theta_i$:

$$x_i \sim P(x;\, \theta_i).$$

In population models, the assumption is that $\theta_i = \theta$ for all $i$. In context-adaptive models, we instead posit that the parameters vary with context:

$$\theta_i = f(c_i) \quad \text{or} \quad \theta_i \sim P(\theta \mid c_i),$$

where $c_i$ captures the relevant covariates or environment for observation $i$. The goal is to estimate either a deterministic function $f$ or a conditional distribution over parameters.

These provide new tools for understanding personalized and context-adaptive systems, but also present new modeling challenges. Estimating a unique $\theta_i$ from a single observation is ill-posed without structural regularization—smoothness, sparsity, shared representations, or latent grouping.

This problem regularly appears in disciplines that rely on observational data, including biology, medicine, finance, language modeling, and the social sciences. As such, developments have come from many diverse research threads. The sample-specific view, where each sample carries its own parameters $\theta_i$, descends from mixed-effects and local-likelihood estimation [1]. Letting parameters vary across discrete groups defined by categorical side information was introduced with mixed-effects models [2]. Letting parameters vary smoothly as a function of an observed covariate $c_i$ was introduced with varying-coefficient models [3]. Extensions to the high-dimensional, structured settings were made in work on time-varying and varying-coefficient network estimation [4,5,6]. As adaptivity becomes more implicit (e.g., via neural networks or black-box inference), new tools will be required to recover, interpret, or constrain the underlying parameter variation.

## Scope and Relation to Prior Work

In this work, we first review methods that use context to guide inference, either by specifying how parameters change with covariates or by learning to adapt behavior implicitly. We begin with classical models that impose explicit structure, such as varying-coefficient models and multi-task learning, and then turn to more flexible approaches like meta-learning and in-context learning with foundation models that fully subsume traditional estimation and inference.

Although these approaches arise from different traditions, they share a common goal: to tailor inference to the local characteristics of each observation or task. We formalize this connection through a theoretical bridge, finding that they rely on strikingly similar estimators both explicitly and in-context. We then use this unified perspective to propose principles for designing, evaluating, and interpreting context-adaptive models.

Along the way, we highlight recurring themes: 1. Complex models often decompose into simpler, context-specific components; 2. Models that accommodate more heterogeneity outperform task-specific predictors by consuming much larger datasets; and 3. Inference-time programmability is a driving force behind broad model adoption. These perspectives offer a unifying lens on recent advances and open new directions for building adaptive, interpretable, and personalized models.

## Related Surveys and Reviews

Several surveys have examined specific aspects of context-adaptive inference, but they have largely remained confined to individual methodological traditions. Classical statistical surveys focus on varying-coefficient models and related structured regression methods. In machine learning, surveys on transfer and meta-learning emphasize task adaptation and shared representations, while recent work on foundation models explores the implicit adaptation capabilities of large pretrained models. Table 1 summarizes the scope and coverage of representative surveys.

| Survey | Topic Focus | Scope | Coverage of Adaptivity | Gap Relative to This Work |
|---|---|---|---|---|
| Statistical Methods with Varying Coefficient Models[6,7] | Varying-coefficient modeling | Classical statistical modeling, with parameters expressed as functions of covariates | Explicit adaptivity: parameters change smoothly with context via (f(c)) | Limited to explicit, parametric formulations; no connection to neural or emergent adaptation |
| A Survey of Deep Meta-Learning[8] | Meta-learning | Neural meta-learning methods for cross-task adaptation | Task-level adaptivity: models learn to generalize quickly across tasks | Focused on task switching; does not integrate explicit parameter modeling or implicit foundation model adaptation |

| Survey | Topic Focus | Scope | Coverage of Adaptivity | Gap Relative to This Work |
|---|---|---|---|---|
| LoRA: Low-Rank Adaptation of Large Language Models[9] | Parameter-efficient adaptation | Adaptation of large pretrained transformer models via low-rank updates while freezing base weights | Implicit adaptivity via parameter-efficient updates, enabling contextual adaptation without full fine-tuning | Strong in efficient adaptation mechanism, but narrow in scope; does not address explicit contextual structure or cross-domain generalization |
| Foundational Models Defining a New Era in Vision: A Survey and Outlook[10] | Vision-based foundation models | Architectures, multimodal integration, prompting, fusion in vision models | Implicit adaptivity in vision contexts, via prompt or fusion mechanisms across visual tasks | Domain-specific focus limits generalization; less discussion on theoretical adaptation across modalities |
| A Comprehensive Survey on Pretrained Foundation Models[11] | Pretrained foundation models | Coverage of models across modalities, training regimes, adaptation and fine-tuning strategies | Implicit adaptivity via representation transfer and generalization across tasks | Broad in scope but does not deeply analyze parameter-level adaptation or explicit–implicit alignment |

Table 1: Representative surveys and key papers covering context-adaptive inference. Most works focus on a single methodological tradition and do not connect explicit and implicit approaches.

While existing surveys have reviewed individual components of this landscape the unifying context-adaptive view remains unexplored. This perspective unifies several disparate research threads: Statistical varying-coefficient models, meta-learning / transfer learning, and foundation-model prompting / retrieval. By situating classical statistical models, modern machine learning methods, and foundation models along a shared spectrum of context-adaptive inference, we are able to highlight common principles and distinctive challenges. The sections that follow trace the historical evolution of statistical methods toward context-adaptive inference, develop the explicit and implicit modeling approaches in detail, outline the core theoretical bridge between these threads, and distill the principles that govern when adaptation helps and how to measure it.

## 2. Evolution of Statistical Modeling

Here we provide a general review of how data collection and statistical methods have co-evolved to accommodate increasing data volume, dimensionality, and multi-modality. In parallel, data heterogeneity becomes increasingly problematic as larger data spaces are sampled, and context-adaptivity becomes increasingly essential. Several statistical concepts essential to context-adaptive modeling appear from disparate research threads, such as hierarchical structure and multi-task learning, and are only later unified in context-adaptive approaches.

We trace this co-evolution through a sequence of data regimes. Each regime broke a homogeneity assumption that its predecessor relied on, and each break provoked a more adaptive response, until parameters themselves had to be treated as functions of context.

## Independent and identically distributed samples

The earliest research data came largely from tightly designed experiments, such as agricultural field trials or psychological studies, which were small in scale and simple in structure. Observations were treated as independent and identically distributed [12], with no dependency between units, so inference targeted a single population-level average or effect. This is the regime in which one global parameter suffices, and every method below can be read as a later attempt to relax it.

Linear models were the foundational tool. The method of least squares, first published by Legendre [13] and later justified by Gauss [14], fits a regression line by minimizing squared deviations,

$$\hat{\beta} = \arg\min_{\beta} \sum_{i=1}^{n} \left(y_i - x_i^\top \beta\right)^2,$$

giving a systematic procedure for estimating a shared parameter vector of coefficients $\beta$ from a set of $n$ i.i.d. observations of paired covariates and outcomes $(x_i, y_i)$. Generalized linear models (GLMs) extended this reach to non-Gaussian outcomes. Nelder and Wedderburn connected the mean of the response to a linear predictor through a link function [15], later formalized into the now-standard compact notation [16],

$$g(\mu_i) = x_i^\top \beta,$$

which recovers logistic regression for binary outcomes, following Berkson's dose-response work and its formalization for binary sequences [17,18], and Poisson regression for counts, all within one framework [15]. Alongside regression, analysis of variance (ANOVA) partitioned total variability into within-group and between-group components and tested group differences through the F ratio [19]. Together these frameworks handled independent data well, but they assumed a single parameter set shared by every unit. As studies grew and observations stopped being independent, that assumption was the first to break.

## Hierarchical data

As research expanded, collected data increasingly carried hierarchical structure. Students are nested within classrooms and schools [20], respondents within populations of subjects [21], and repeated measurements within individuals over time. Such data are no longer independent: they carry systematic between-group differences and within-individual correlation. This regime forced the first genuine departure from a single shared parameter, letting parameters vary across groups.

Work on hierarchical dependence began with linear models combining fixed and random effects to estimate genetic parameters [2], partitioning variability into systematic and random components. Recognition that comparative rate estimation depends on the composition of sampled subgroups reinforced the point that heterogeneity in source populations affects inference [22]. Restricted maximum likelihood improved variance-component inference in unbalanced settings [23], enabling the linear mixed-effects model (LMM) that pairs population-level fixed effects with group-specific random deviations [24],

$$y = X\beta + Zu + \varepsilon, \quad u \sim N(0, G), \quad \varepsilon \sim N(0, R).$$

Here $X$ and $Z$ are the fixed- and random-effects design matrices, $\beta$ the population-level fixed effects, $u$ the group-specific random effects with covariance $G$, and $\varepsilon$ the residual error with covariance $R$.

Extensions carried this beyond Gaussian responses: practical estimation for generalized linear mixed models applied link functions to clustered binary, count, or categorical outcomes [25], with penalized quasi-likelihood making application feasible in medical and biological settings [26]. In parallel, Bayesian hierarchical modeling linked parameters across groups through priors [27],

$$y_{ij} \mid \theta_j \sim p(y_{ij} \mid \theta_j), \quad \theta_j \sim p(\theta_j \mid \phi), \quad \phi \sim p(\phi),$$

where $y_{ij}$ is the $i$-th observation in group $j$, $\theta_j$ the group-specific parameters, and $\phi$ the shared hyperparameters that couple groups through the prior.

though it became practical only once Gibbs sampling [28] and later Markov chain Monte Carlo methods [29] supplied the needed computation. These developments let parameters be indexed by group, the first crack toward $\theta_i = f(c_i)$. The variation, however, was still tied to discrete, pre-specified groupings rather than to arbitrary context.

## Functional and high-dimensional data

Through the 1980s and 1990s, observations increasingly arrived as entire curves or functions, such as time series, spectra, or images, rather than fixed-length vectors. Functional data analysis treats each observation as a smooth function $x_i(t)$ over a continuum and represents it with basis expansions [30]. This continuous view demanded more flexible regression, and generalized additive models let each predictor act through an estimated smooth function while retaining interpretability [31],

$$y_i = \alpha + \sum_{j=1}^{p} f_j(x_{ij}) + \varepsilon_i,$$

where $\alpha$ is an intercept, each $f_j$ is an estimated smooth function of the $j$-th of $p$ predictors $x_{ij}$, and $\varepsilon_i$ is residual error. In this case the data, rather than a fixed linear form, determine the shape of each effect. As dimensionality kept growing, for example with genomic or imaging data, the burden shifted from specifying features to learning them. Representation learning generalizes classical dimension reduction such as PCA to nonlinear, data-driven embeddings [32], with neural autoencoders as the prototypical case: an encoder maps each input to a low-dimensional code and a decoder reconstructs it, so the model adapts its own feature extraction to complex data. In this regime, heterogeneity no longer lives in a hand-specified design but in a learned representation, which raises the question of how that representation should itself depend on context.

## Heterogeneous tasks and scarce data

As computation became cheap and universal, models could be estimated faster than data could be collected for each task. To accommodate, statistical learning shifted toward sharing information across problems, especially when each problem individually carried little data. Multi-task learning jointly models related tasks so that information useful for one aids the others [33],

$$\min_{W} \sum_{t=1}^{T} \sum_{i=1}^{n_t} \ell\big(y_i^t, f(x_i^t; w^t)\big) + \lambda\, \Omega(W),$$

where $W$ collects the per-task parameter vectors $w^t$, $\ell$ is a per-task loss over the $n_t$ samples of each of $T$ tasks, $\lambda$ sets the regularization strength, and $\Omega(W)$ enforces shared structure. Transfer learning extends the idea across domains, adapting a model trained on a large source to a target with few labels [34], as when an ImageNet-pretrained network is fine-tuned on a small medical imaging set. Distribution shift between training and deployment introduced a further challenge: under covariate shift the input distribution changes while the conditional stays fixed, correctable by importance weighting, and more general domain adaptation handles shifts in both the input and the conditional through techniques such as kernel mean matching or adversarial domain-invariant representations. Pushing to the extreme of many tasks with only a handful of examples each yields few-shot learning [35], where metric-based methods embed inputs so that same-class samples cluster [36,37] and meta-learning trains across simulated tasks so a model learns to adapt quickly. These frameworks confront scarcity by transferring inductive bias across tasks, making the task itself a form of context and adaptation an explicit goal, a thread the implicit-adaptivity section develops in full.

## Online and interactive data

In the Internet era, data increasingly arrived as streams shaped by interaction and feedback [38], breaking the assumption that a fixed distribution can be sampled once. Online learning addresses this through the online convex optimization framework, where a learner updates a decision each round against an adversarially revealed loss and is judged by its regret relative to the best fixed decision in hindsight [39,40]; online gradient descent achieves sublinear regret in this setting. Because real streams are non-stationary, the input-output relationship can itself drift over time [41], motivating drift-detection and adaptive-windowing methods that reset or reweight the model when the error distribution shifts [42,43,44].

In this regime, context becomes something the system must act on. The multi-armed bandit formalizes the exploration-exploitation trade-off under partial feedback [45], with classic strategies such as $\epsilon$-greedy [46] and upper-confidence-bound selection [47]. Their key limitation, ignoring side information, leads directly to contextual bandits, where the agent observes a context $x_t$ each round and learns a policy mapping context to action. A representative algorithm, LinUCB [48], assumes a linear context-reward relationship and selects

$$a_t = \arg\max_{a \in \mathcal{A}} \left( x_t^\top \hat{\theta}_a + \alpha \sqrt{x_t^\top A_a^{-1} x_t} \right),$$

where $\mathcal{A}$ is the action set, $\hat{\theta}_a$ the estimated reward parameters for arm $a$, $A_a$ its design matrix, and $\alpha$ controls the strength of the exploration bonus. This setup personalizes decisions to the observed environment, as in news recommendation and adaptive experimentation. When actions also influence future contexts, reinforcement learning generalizes this to sequential decisions through Markov decision processes [49], with value-based [50] and policy-based [51,52] solutions enabling applications from game-playing to robotics [53]. Across this progression, adaptivity to context and feedback moves from a modeling convenience to the central object.

## Multimodal data and large-scale pretraining

Digitalization brought data spanning several modalities at once, including images [54], audio [55], and text [56]. These sources are not only high-dimensional but structurally distinct: text is discrete and symbolic, images are spatially organized, and audio is a continuous waveform. Modeling them together requires establishing correspondences across modalities, building shared latent spaces, and adjusting inter-modal dependencies by task. Representation learning supplies the shared latent space by mapping heterogeneous inputs into a common representation, and recent theory frames this adaptability directly in terms of context: the Contexture Theory casts representation learning as approximating the expectation operator induced by a context variable, unifying supervised, self-supervised, and manifold learning under one spectral view and showing that constructing informative context can be more efficient than merely scaling capacity [57].

Scaling this further, web-scale collection produced massive, weakly supervised, cross-domain corpora, such as large text collections [58] and image-text alignment data [59], gathered under open conditions without explicit task labels. The modeling goal shifts from fitting one distribution to extracting patterns that remain stable across heterogeneous contexts, which is the premise of foundation models: general-purpose systems pretrained at scale and adapted to downstream tasks with little additional supervision [60]. Model performance follows predictable power laws in parameter count, dataset size, and compute, leading to the large-scale pretraining regime common today [61,62]. Pretraining is often carried out through masked-language modeling or causal language modeling. These objectives use a variable mask to corrupt the input data, and require the model to reconstruct the corrupted tokens given the uncorrupted tokens. This gives a joint distribution over the entire sample

$$\log p(x) = \sum_{t=1}^{T} \log p(x_t \mid x_{<t}),$$

where $x_t$ is the $t$-th token and $x_{<t}$ the preceding tokens of a length-$T$ sequence, and allows context to be flexibly specified. This allows adaptation to occur over a wide range of contexts at inference time, without weight updates: given a prompt of examples $(x_1, y_1), \ldots, (x_k, y_k)$ and a query $x_{k+1}$, the model predicts

$$\hat{y}_{k+1} = \arg\max_{y_{k+1}} p(y_{k+1} \mid x_{k+1}, y_k, x_k, \ldots y_1, x_1),$$

with the same fixed pretrained parameters $\theta^{\star}$. This emergent capability, in-context learning, is the basis for the implicit context-adaptive view explored in subsequent sections.

### From heterogeneity to context

This history is a progressive erosion of the i.i.d. baseline. Each regime adds a source of heterogeneity that a single global parameter could not accommodate: dependence across groups, structure in high dimensions, variation across tasks, drift and feedback over time, and finally differences across modalities and domains at web scale. The recurring response was to let parameters vary, first across known groups, then smoothly across covariates, then across tasks and data modalities. The following sections explore this progression through the $\theta_i = f(c_i)$ formalism introduced earlier, where models are learned functions of arbitrary context.

## 3. Explicit Adaptivity: Structured Estimation of $f(c)$

To better capture heterogeneity across contexts, some recent approaches model parameters as explicit functions of observed context, formalized as $\theta_i = f(c_i)$, where $f$ maps each context to a sample-specific parameter [3].

A familiar example of explicit adaptivity is multi-task learning, where context is defined by a task flag or identity. Traditional multi-task learning (left) assigns each task its own head on top of shared representations, while context-flagged models (right) pass task identity directly as an input, enabling richer parameter sharing. This illustrates how explicit conditioning on context variables can unify tasks within a single model and provides an intuitive entry point to more general forms of explicit adaptivity (Figure 3).

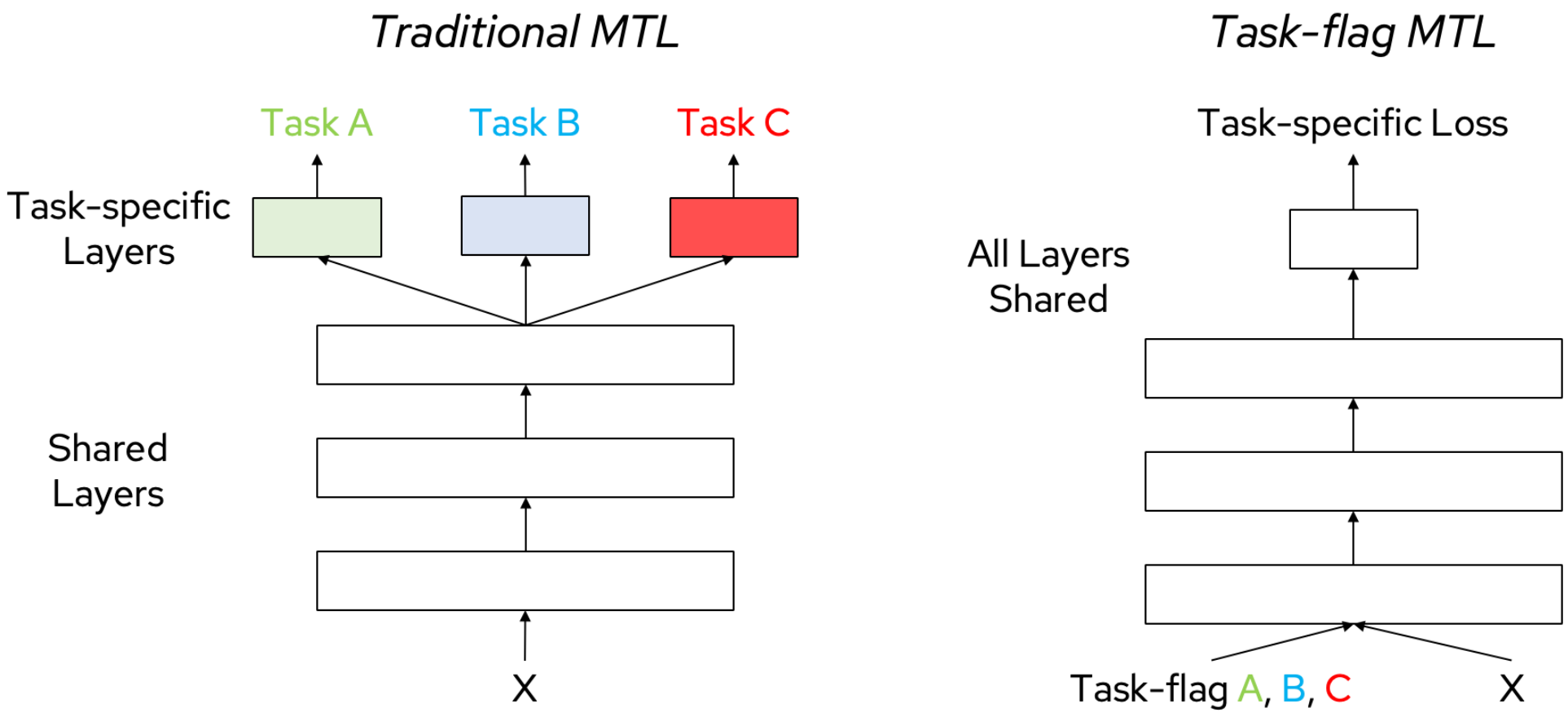

**Figure 3:** Multi-task learning as explicit adaptivity. In traditional MTL (left), each task has its own head on top of shared layers. In context-flagged models (right), the task identity is provided as an input, enabling a shared model to adapt across tasks.

The canonical formalism for explicit adaptivity is the varying-coefficient model (VCM), which writes each regression coefficient as a function of context [3,63]:

$$y_i = \sum_{j=1}^{p} \beta_j(c_i)\, x_{ij} + \varepsilon_i.$$

where each coefficient $\beta_j(c_i)$ is a function of the context $c_i$, $x_{ij}$ is the $j$-th of $p$ covariates, and $\varepsilon_i$ is residual error. Every method in this section is, at its core, a different way to estimate the maps $\beta_j(\cdot)$, or more generally $f(\cdot)$. We organize them along an axis of increasingly sophisticated statistical and machine-learning concepts. This progression traces a spectrum of assumptions about how parameters relate to context, from none to fully learned:

- Global models: $\theta_i = \theta$ for all $i$.
- Grouped models: $\theta_i = \theta_c$ for a finite set of groups.
- Smooth models: $\theta_i = f(c_i)$, with $f$ continuous or low-complexity.
- Latent models: $\theta_i \sim P(\theta \mid c_i)$, with $f$ learned implicitly.

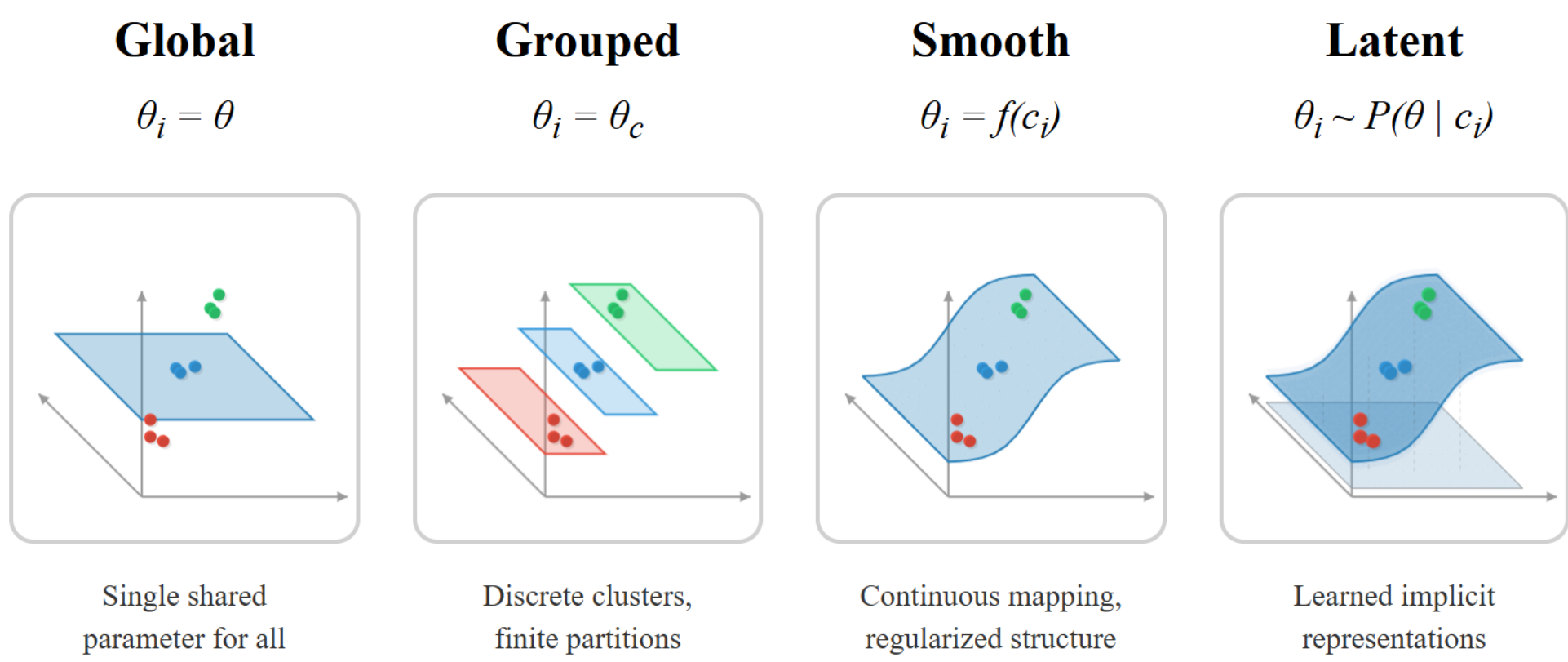

**Figure 4:** A spectrum of context awareness in modeling, showing global, grouped, smooth, and latent models

Each step forward increases the power of context-specific, personalized inference by borrowing strength from related samples and groups. Each step also lowers the amount of data that must be collected for any single context, because information flows in from neighboring contexts rather than being estimated in isolation.

The sections below start from subgroups that share nothing, through mechanisms that share progressively more, to fully learned functions of context: kernel localization that dissolves boundaries entirely, structured parametric maps that learn which context features matter, and finally contextualized models that learn arbitrary context dependencies. Running alongside this is a parallel case in which the models being adapted become increasingly sophisticated, going from linear models to graphical models whose structure, and not only its values, follows context.

## Independent Subgroups

The most basic response to heterogeneity is to split the data into subgroups and fit each one on its own. Conditional and clustered models define groups by hand, for example by sex or site, or by unsupervised clustering, and estimate a separate parameter vector within each group with no sharing between them,

$$\{\widehat{\theta}_0, \ldots, \widehat{\theta}_C\} = \arg\max_{\theta_0,\ldots,\theta_C} \sum_{c\in\mathcal{C}} \ell(X_c; \theta_c),$$

where $\ell(X_c; \theta_c)$ is the log-likelihood of $\theta_c$ on the samples assigned to group $c$. This removes the bias of a single global fit, but it is the most data-hungry option available: each group is estimated only from its own samples, so small groups are estimated poorly and unseen groups not at all. Every later step can be read as a way to relax this isolation, buying more reliable per-group estimates from less per-group data.

## Sharing Across Groups

The first improvement keeps discrete groups but couples their estimates. Distance-regularized estimation asks that observations with similar contexts have similar parameters, penalizing differences in $\theta_i$ in proportion to context distance,

$$\{\widehat{\theta}_0, \ldots, \widehat{\theta}_N\} = \arg\max_{\theta_0,\ldots,\theta_N} \left( \sum_i \ell(x_i; \theta_i) - \sum_{i,j} \frac{\|\theta_i - \theta_j\|}{D(c_i, c_j)} \right),$$

where $D(c_i, c_j)$ is a distance metric between contexts, while fused-lasso and total-variation penalties shrink differences between adjacent groups so that estimates borrow strength from their neighbors [64]. The effect is to lower the effective data requirement per group: a sparsely observed context inherits information from well-observed relatives rather than standing alone.

The same idea carries over from regressors to network estimators. The graphical-model lineage established that a parameter can be an estimated network: zeros in the inverse covariance matrix encode conditional independencies [65,66], and neighborhood selection and the graphical lasso recover sparse structure directly from data [67,68], with later extensions estimating this structure when each node carries a vector of attributes rather than a scalar [69].

Most of the graph estimators above use regularizers to enforce graph structure. A direct extension is to regularize networks across groups, which couples the estimation tasks and enables information sharing. Guo et al. jointly estimate several graphical models, encouraging sparsity within each while borrowing strength across related groups [70], and the Joint Graphical Lasso balances shared structure against group-specific edges across populations [71]. Bayesian formulations achieve the same pooling through priors rather than penalties: lattice and Markov-random-field spike-and-slab priors learn when edges should be shared across neighboring sites or sample groups, and quantify how similar the resulting networks are [72,73].

## Separating Subgroups

The methods so far encourage parameters for similar contexts to converge. A converse design goal sometimes matters more: keeping the models for distinct subgroups separated so that minority subgroups do not collapse into majority patterns. This negative information sharing is typically implemented by learning representations that disentangle subgroup structure, connecting statistical partitioning with adversarial or contrastive objectives [74].

## Learning the Group Boundaries

Thus far group-based modeling has required knowing which groups to model ahead of estimation. The next step lets the data decide where the boundaries fall. When samples are ordered along a context such as time, a total-variation penalty on successive parameters yields piecewise-constant estimates whose breakpoints are inferred rather than fixed,

$$\{\widehat{\theta}_0, \ldots, \widehat{\theta}_N\} = \arg \max_{\theta_0, \ldots, \theta_N} \left( \sum_i \ell(x_i; \theta_i) + \lambda \sum_{i=2}^{N} \|\theta_i - \theta_{i-1}\| \right),$$

with the penalty strength $\lambda$ trading the number of breaks against fit. Model-based recursive partitioning fits a parametric model, tests its coefficients for instability across candidate context variables, and splits on the variable with the strongest instability before recursing, so the partition is chosen by the data rather than pre-specified [75]; toolkits such as partykit fit a parametric model within each leaf [76]. In the network case the same move recovers where structure changes. Varying-coefficient varying-structure (VCVS) graphical models treats both the edge set and the edge weights as functions of context, and when structure changes abruptly a temporally smoothed penalty recovers the changepoints together with the precision matrix on each block, the first such estimators shown to be sparsistent with established convergence rates [6,77]. TREEGL extends this to a tree of networks that switch along a branching biological lineage, borrowing strength between a parent cell type and its descendants while exposing the edges that change at each division [78,79].

Viewing a partition as explicit routing clarifies what these splits accomplish. Each split of the context space sends samples to a distinct parameter vector, so the boundaries encode exactly where parameters are shared and where they are separated. Hierarchical partitions capture heterogeneity at two levels, sample-level variation within a context and task-level switching across contexts, which connects partition-based models to multi-task learning (Figure 5). This explicit routing is the structured counterpart of the implicit routing performed by mixture-of-experts and attention layers in the next section, where the same shared-versus-separated decision is made inside the network rather than by an overt split.

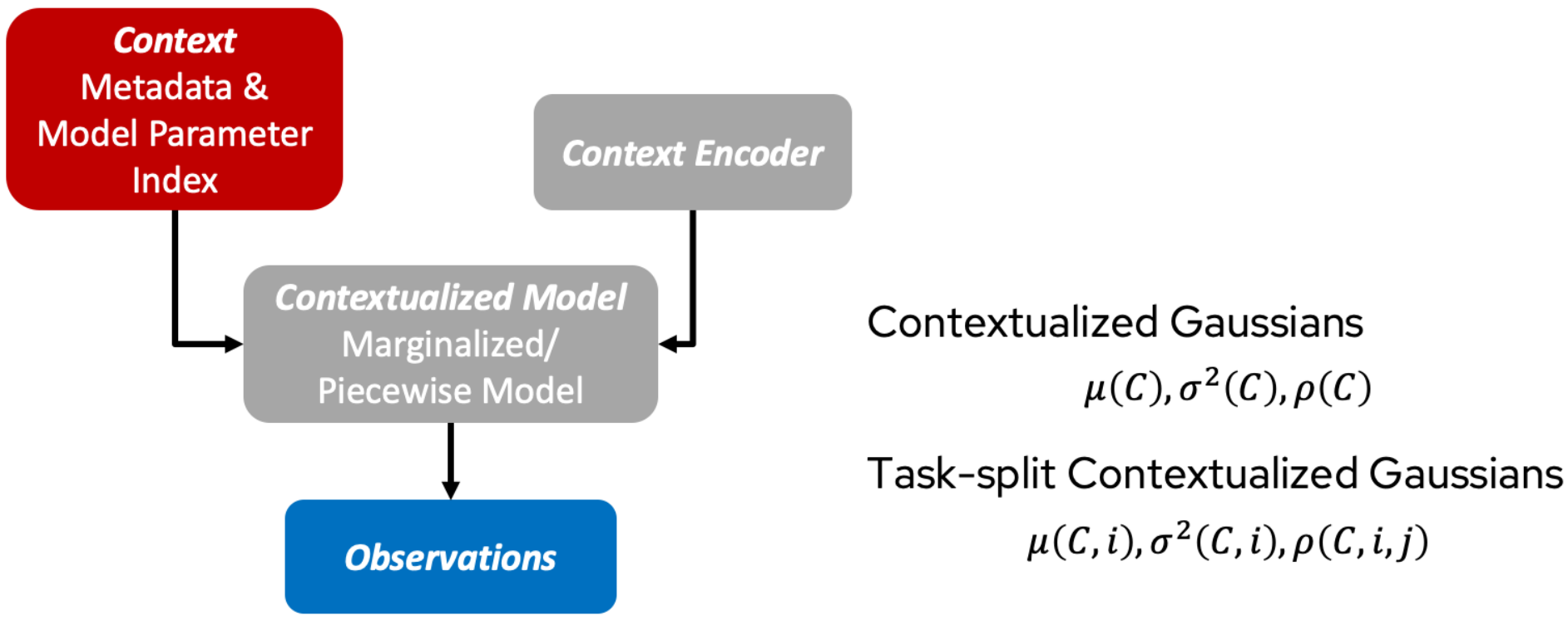

**Figure 5:** Hierarchical splits of context enable multi-level adaptivity. Explicit adaptivity can partition the context space into piecewise models, with parameters indexed both by context $c$ and task identity $(i, j)$. Such splits allow sample-level heterogeneity to be captured within contexts, while high-level partitions mimic task boundaries and enable task switching.

## Dissolving the Boundaries with Kernels

Partitions, however they are drawn, still impose hard boundaries. Kernel and locally weighted methods remove them, estimating a separate model at each query context from a similarity-weighted neighborhood of samples. This is the classical varying-coefficient setting, where coefficients are smooth functions of a low-dimensional context estimated with kernel smoothing, local polynomials, or penalized splines [80,81]. For a query context $c^*$, a semiparametric VCM solves

$$\widehat{\theta}(c^*) = \arg\max_{\theta} \sum_{i=1}^{n} K_\lambda(c_i, c^*)\, \ell(x_i; \theta),$$

where $K_\lambda$ measures similarity between contexts and $\ell$ is the per-sample log-likelihood, so prediction at $c^*$ is a similarity-weighted combination of nearby observations. Because every sample contributes in proportion to its context similarity, no single group needs to be large; the neighborhood supplies the data. Reproducing-kernel methods extend this to higher dimensions while retaining guarantees, for example penalized RKHS estimators for partially varying-coefficient models that separate constant from smoothly varying effects with minimax prediction rates and consistent structure selection [82], and the same machinery underlies generalized additive models. This locally weighted view is also the thread a later section follows to show that transformers performing in-context learning realize the same estimator, with a learned attention kernel taking the place of $K_\lambda$.

Similarity can be defined over a topology rather than a continuous covariate. Spatially varying-coefficient models let local effects change gradually across adjacent regions [83,84], the network varying-coefficient model learns latent node positions and coefficient functions on a graph [85], and Laplacian or nested-group penalties encode smoothness over temporal, hierarchical, or multilevel structure. The network-valued case appears here too: kernel reweighting and total-variation penalties estimate a separate network at each point along a context axis, as in TESLA and related kernel-reweighted and time-varying network estimators for rewiring gene-regulatory and political networks [4,5,64,86,87,88], and covariate-dependent Bayesian graph learning lets network structure vary smoothly with observed covariates through a dual spike-and-slab prior that selects at node, covariate, and local levels [89].

## Structured Parametric Maps

Kernel methods treat the context dimensions symmetrically through the similarity metric. The next step learns which dimensions matter. Structured parametric VCMs impose form on $f$. The simplest is a linear map $\theta_i = Ac_i$, estimated by

$$\widehat{A} = \arg\max_A \sum_i \ell(x_i; Ac_i),$$

where sparsity and group penalties such as the L1 lasso can be imposed on the coefficient function to identify the context features driving adaptation [90]. Tree-based ensembles extend this to tabular and mixed-type data: Tree Boosted Varying-Coefficient Models estimate context-dependent coefficients with gradient-boosted trees, balancing flexibility, accuracy, and interpretability while remaining easier to tune than deep networks [91]; cyclic gradient boosting adds dimension-wise early stopping and feature-importance measures [92]; and VCBART embeds Bayesian Additive Regression Trees into the varying-coefficient framework, estimating complex effect modifiers with coherent uncertainty quantification and good scaling in high dimensions [93]. What these share is a readout of context relevance, through split statistics, feature importances, or posterior inclusion, that the kernel view does not provide.

## From Composition to Learned Functions

The jump from parametric to nonparametric adaptivity is straightforward. If we fit simple parametric models within each context, for observed contexts $c$ or latent subcontexts $Z$, and then aggregate across contexts, the resulting conditional

$$P(Y \mid X, C) \;=\; \int P(Y \mid X, C, Z)\, dP(Z \mid C)$$

can display rich, multimodal behavior that looks nonparametric. Global flexibility can emerge from compositional, context-specific parametrics. When component families are identifiable or suitably regularized and the context-to-mixture map is constrained by smoothness, total variation, or sparsity over $c$, the aggregate model remains estimable and interpretable while avoiding overflexible, ill-posed mixtures (Figure 6).

Nonparametric inference from context-adaptive parameters

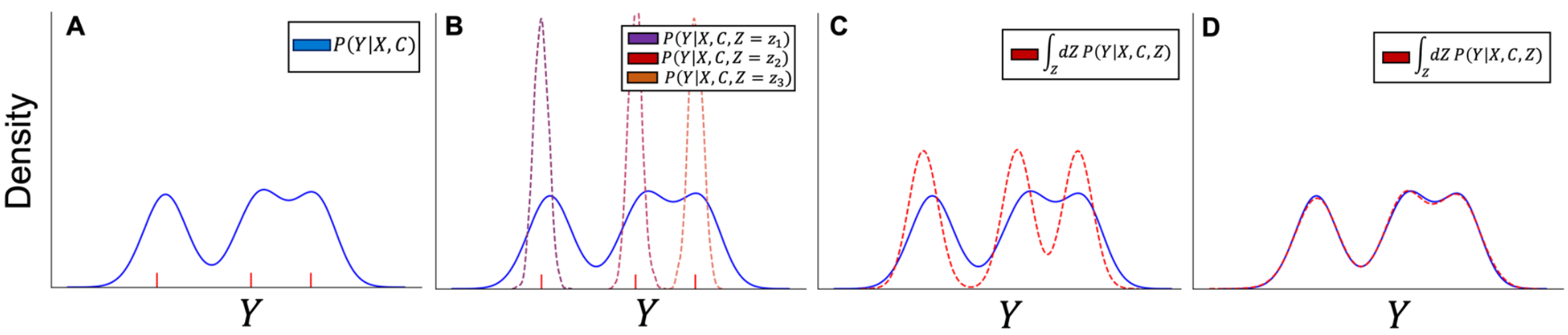

**Figure 6:** Compositional inference: nonparametric flexibility from parametric context-specific models. (A) Overall conditional $P(Y \mid X, C)$. (B) Context-specific components $P(Y \mid X, C, Z = z_i)$ for latent subgroups $Z$. (C) Recombination via marginalization $\int_Z P(Y \mid X, C, Z)$. (D) Aggregated distribution showing how structured parametric pieces yield multimodal, nonparametric-like behavior.

This perspective motivates flexible function approximators: trees and neural networks can be read as learning either the context-to-mixture weights or the local parametric maps. The final step lets that function be arbitrary.

## Contextualized Models: Arbitrary Functions of Context

For contexts defined by high-dimensional or unstructured features such as images, text, or sequences, deep neural networks approximate $f(c)$ without committing to a fixed basis, partition, or similarity metric. A network can consume context as input and let its nonlinear layers induce context-dependent behavior, the pattern that reappears for implicit models in the next section; it can generate the parameters of another model directly, as in hypernetworks, which give an explicit deep-net realization of $\theta = f(c)$ [94]; or it can modulate a shared backbone through feature-wise affine transformations supplied by context, as in FiLM [95]. Taking this to its limit yields contextualized models, which treat $f$ as an unrestricted map learned end-to-end and estimate it directly,

$$\widehat{f} = \arg\max_{f \in \mathcal{F}} \sum_i \ell(x_i; f(c_i)),$$

capturing complex functions of context including the feature interactions that the structured maps above cannot express [96]. A single deep encoder reads a sample's context and emits the parameters of its downstream model, so estimation is amortized across contexts: the cost is paid once during training, and inference for a new context is a forward pass that needs no per-group data at all. The formulation spans model types and domains, including personalized disease models [97,98,99], heterogeneous treatment effects [100,101,102], contextual feature selection and explainability [103,104], and drug development [105], with standard implementations in the contextualized.ml package [106]. The network-valued thread reaches the same endpoint: personalized regression and Bayesian edge-regression models learn a map from a sample's covariates or latent similarity to its own network, recovering subject-specific structure rather than a shared group label [107,108]. Because the encoder infers parameters from context alone, it is the explicit object the next section reaches back to when it interprets an amortized context encoder as the bridge to in-context learning.

## Beyond Covariates: What Serves as Context

Context need not be a covariate, a task identifier, or a position in a sequence. Any signal the model can condition on can serve as context, and a useful example is the pattern of missing measurements itself. Combining real-world datasets is complicated by inconsistent measurement: different cohorts or institutions collect different subsets of features, so naive pooling yields a sparse, unbalanced feature matrix, while discarding incomplete samples wastes data. Context-adaptive models resolve this by treating measurement sparsity as context. Rather than ignoring missingness, the model adjusts its parameterization to which features are observed, so that each measurement policy, whether labs-only, vitals-only, or multimodal, defines a context and information is shared across policies while their differences are respected. This reframes missingness from a nuisance into structured signal that encodes which sources of evidence are available and how they should be combined, an idea also pursued in multimodal learning frameworks that handle missing modalities [109]. By conditioning on measurement availability, a model learns from fewer individuals with more heterogeneous features (Figure 7). The metrics and stress tests specific to missingness-as-context are collected with the shared evaluation principles later in the review.

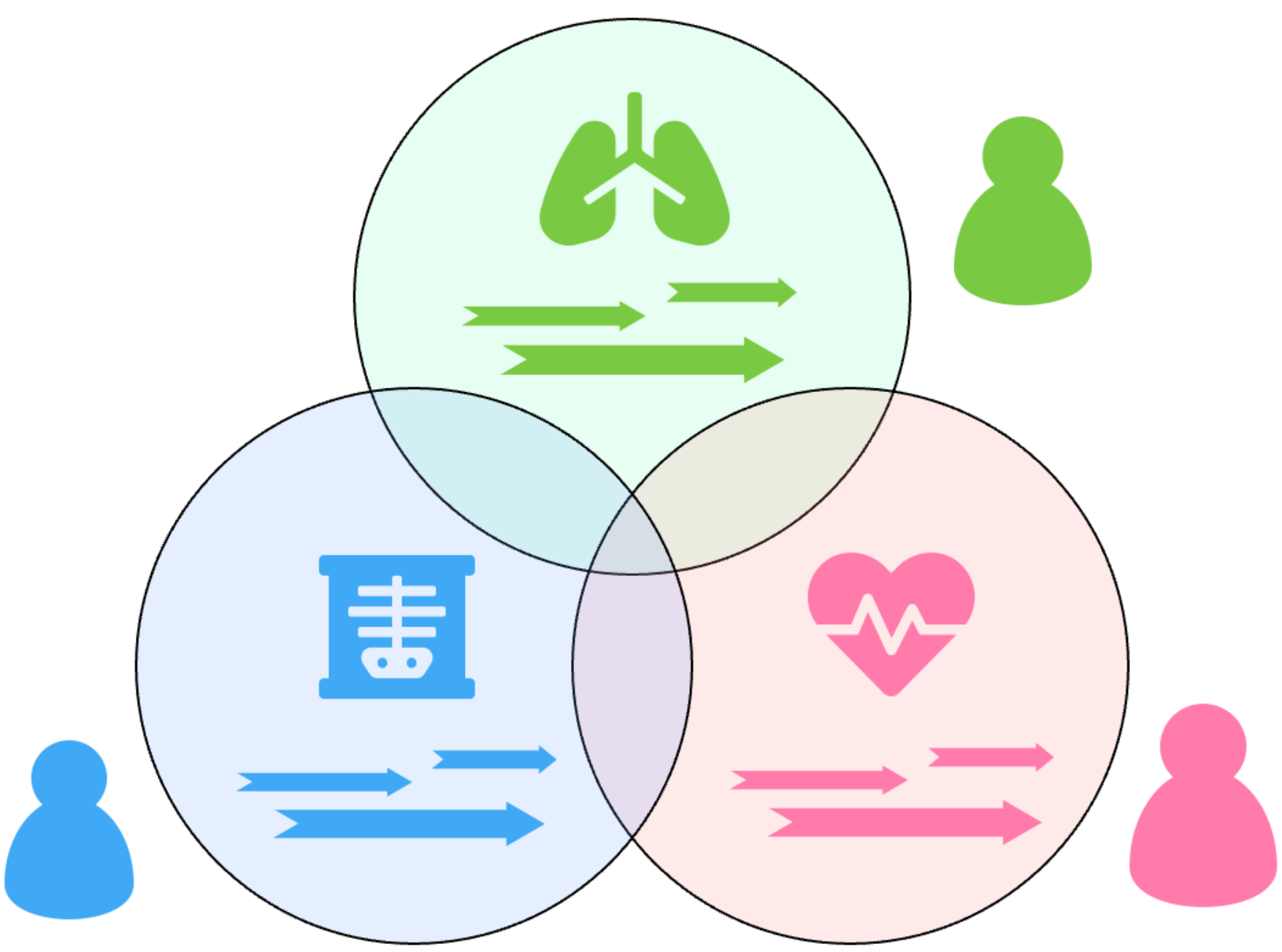

**Figure 7:** Patterns of missingness as context. Each dataset (e.g., cohort with labs, cohort with vitals, cohort with imaging) provides a different subset of measurements. Context-adaptive models allow integration by conditioning on measurement availability, enabling learning from fewer samples with more heterogeneous features.

## Key Theoretical Advances

The guarantees available for these methods reflect the assumptions each one makes, and they track the progression above.

### Partitions and learned boundaries

For grouped and partition-based estimators, change-point analysis and total-variation regularization establish when abrupt parameter changes can be recovered. Under suitable sparsity and signal-strength conditions, fused-lasso and total-variation penalties recover both the location of the changepoints and the parameters on each piecewise-constant block [64].

### Kernel and smooth estimation

For kernel smoothing, local polynomial estimation, and penalized splines, convergence rates and efficiency are well characterized. Under standard regularity conditions these estimators achieve minimax-optimal rates for function estimation in moderate dimensions [3], and Lu, Zhang, and Zhu established consistency and asymptotic normality for penalized spline estimators given a sufficient number of knots and appropriate penalties, enabling valid inference through confidence intervals and hypothesis tests [110].

### Structure-varying estimation

When the support itself varies with context, analysis centers on identifiability, sparsistency, and consistent recovery of the changing structure. The VCVS estimators of Kolar and Xing were the first shown to be sparsistent with established convergence rates when the network changes abruptly over time [6,77], and for networked coefficients non-asymptotic error bounds show that consistency is attainable when the underlying graph topology is sufficiently connected [85].

### Learned and high-capacity models

High-capacity approximators raise harder questions. In high-dimensional and sparse settings, oracle inequalities and penalized-likelihood theory give conditions for consistent variable selection and accurate estimation, including for boosting-based estimators. For neural-network realizations of $f(c)$, generalization and identifiability under non-convex optimization remain only partly understood and are an active frontier for both the statistics and machine learning communities.

## Future Directions

One direction integrates varying-coefficient models with foundation models from language and vision. Using pretrained embeddings as the context $c_i$ lets these models absorb large amounts of prior knowledge and extend to multimodal and unstructured sources, and the principled combination of cross-modal contexts, bringing text, images, and structured covariates into a single framework, remains an open problem.

Interpretability and visualization for high-dimensional or black-box coefficient functions are equally important. Tools that let users understand and trust estimated coefficient surfaces are a prerequisite for adoption in sensitive areas such as healthcare and policy.

Finally, the gap between methodological innovation and practical deployment remains. Many capable VCM variants exist, but adoption is often limited by the availability of software and the clarity of methodological guidance [63]. Continued investment in usable implementations, open-source libraries, and empirical benchmarks will broaden adoption and impact. Explicit adaptivity now spans this full progression, from independent subgroups to amortized deep encoders; the principles for evaluating and deploying any point on it, shared with the implicit methods of the next section, are the subject of a later chapter.

# 4. Implicit Adaptivity: Emergent Contextualization in Complex Models

Traditional models often describe how parameters change by directly specifying a function of context, for example through expressions like $\theta_i = f(c_i)$, where the link between context $c_i$ and parameters $\theta_i$ is fully explicit. In contrast, many modern machine learning systems adapt in fundamentally different ways. Large neural network architectures, particularly foundation models that are now central to state-of-the-art AI research [60], exhibit forms of adaptation that do not arise from any predefined mapping. Instead, their flexibility emerges from the interaction between model structure and the breadth of training data—an effect we refer to as implicit adaptivity. They show a capacity for adaptation that does not arise from any predefined mapping. Instead, their flexibility emerges naturally from the structure of the model and the breadth of the data seen during training. This phenomenon is known as implicit adaptivity. This emergent phenomenon, referred to as implicit adaptivity, highlights how learning and inference can become intertwined within the model itself. Attention

Unlike explicit approaches, implicit adaptivity does not depend on directly mapping context to model parameters, nor does it always require context to be formally defined. Such models, by training on large and diverse datasets, internalize broad statistical regularities. As a result, they often display context-sensitive behavior at inference time, even when the notion of context is only implicit or distributed across the input. This capacity for emergent adaptation is especially prominent in foundation models, which can generalize to new tasks and domains without parameter updates, relying solely on the information provided within the input or prompt.

In this section, we offer a systematic review of the mechanisms underlying implicit adaptation. We first discuss the core architectural principles that support context-aware computation in neural networks. Next, we examine how meta-learning frameworks deliberately promote adaptation across diverse tasks. Finally, we focus on the advanced phenomenon of in-context learning in foundation models, which highlights the frontiers of implicit adaptivity in modern machine learning. Through this progression, we aim to clarify the foundations and significance of implicit adaptivity for current and future AI systems.

## Foundations of Implicit Adaptation

The capacity for implicit adaptation does not originate from a single mechanism, but reflects a range of capabilities grounded in fundamental principles of neural network design. Unlike approaches that adjust parameters by directly mapping context to coefficients, implicit adaptation emerges from the way information is processed within a model, even when the global parameters remain fixed. To provide a basis for understanding more advanced forms of adaptation, such as in-context learning, this section reviews the architectural components that enable context-aware computation. We begin with simple context-as-input models and then discuss the more dynamic forms of conditioning enabled by attention mechanisms.

### Architectural Conditioning via Context Inputs

In contrast to explicit parameter mapping, the simplest route to implicit adaptation is to feed context directly as part of the input. The simplest form of implicit adaptation appears in neural network models that directly incorporate context as part of their input. In models written as $y_i = g([x_i, c_i]; \Phi)$, context features $c_i$ are concatenated with the primary features $x_i$, and the mapping $g$ is determined by a single set of fixed global weights $\Phi$. Even though these parameters do not change during inference, the network's nonlinear structure allows it to capture complex interactions. As a result, the relationship between $x_i$ and $y_i$ can vary depending on the specific value of $c_i$.

This basic yet powerful principle is central to many conditional prediction tasks. For example, personalized recommendation systems often combine a user embedding (as context) with item features to predict ratings. Similarly, in multi-task learning frameworks using task-flags (Figure 3), shared networks learn representations conditioned on task or environment identifiers, which allows a single model to solve multiple related problems [111].

### Interaction Effects and Attention Mechanisms

Modern architectures go beyond simple input concatenation by introducing interaction layers that support richer context dependence. These can include feature-wise multiplications, gating modules, or context-dependent normalization. Among these innovations, the attention mechanism stands out as the foundation of the Transformer architecture [112].

Attention allows a model to assign varying degrees of importance to different parts of an input sequence, depending on the overall context. In the self-attention mechanism, each element in a sequence computes a set of query, key, and value vectors. The model then evaluates the relevance of each element to every other element, and these relevance scores determine a weighted sum of the value vectors. This process enables the model to focus on the most relevant contextual information for each step in computation. The ability to adapt processing dynamically in this way is not dictated by explicit parameter functions, but emerges from the network's internal organization. By enabling dynamic, input-dependent weighting, attention supports context-aware computation without altering global parameters, thereby setting the stage for advanced on-the-fly adaptation such as in-context learning.

## Amortized Inference, Meta-Learning, and Context Encoding

Moving beyond fixed architectures that implicitly adapt, another family of methods deliberately trains models to become efficient learners. These approaches, broadly termed meta-learning or "learning to learn," distribute the cost of adaptation across a diverse training phase. As a result, models can make rapid, task-specific adjustments during inference. Rather than focusing on solving a single problem, these methods train models to learn the process of problem-solving itself. This perspective provides an important conceptual foundation for understanding the in-context learning capabilities of foundation models.

### Amortized Inference

Amortized inference represents a more systematic form of implicit adaptation. In this setting, a model learns a reusable function that enables rapid inference for new data points, effectively distributing the computational cost over the training phase. In traditional Bayesian inference, calculating the posterior distribution for each new data point is computationally demanding. Amortized inference addresses this challenge by training an "inference network" to approximate these calculations. A classic example is the encoder in a Variational Autoencoder (VAE), which is optimized to map high-dimensional observations directly to the parameters, such as mean and variance, of an approximate posterior distribution over a latent space [113]. The inference network thus learns a complex, black-box mapping from the data context to distributional parameters. Once learned, this mapping can be efficiently applied to any new input at test time, providing a fast feed-forward approximation to a traditionally costly inference process.

### Meta-Learning: Learning to Learn

Meta-learning builds upon these ideas by training models on a broad distribution of related tasks. The explicit goal is to enable efficient adaptation to new tasks. Instead of optimizing performance for any single task, meta-learning focuses on developing a transferable adaptation strategy or a parameter initialization that supports rapid learning in novel settings [114].

Gradient-based meta-learning frameworks such as Model-Agnostic Meta-Learning (MAML) illustrate this principle. In these frameworks, the model discovers a set of initial parameters that can be quickly adapted to a new task with only a small number of gradient updates [115]. Training proceeds in a nested loop: the inner loop simulates adaptation to individual tasks, while the outer loop updates the initial parameters to improve adaptability across tasks. As a result, the capacity for adaptation becomes encoded in the meta-learned parameters themselves. When confronted with a new task at inference, the model can rapidly achieve strong performance using just a few examples, without the need for a hand-crafted mapping from context to parameters. In this view, the capacity to adapt becomes encoded in the meta-learned parameters themselves, enabling rapid generalization from few examples without a hand-crafted map from context to coefficients and standing in clear contrast to explicit approaches.

## Amortized Estimation: Context Encoders

The contextualized models at the end of the previous section already learned a context encoder, a map from a task's context to a set of task-specific parameters. The result is an amortized estimator, which predicts the parameters of the downstream model that would have been produced by a classical estimator if sufficient data were collected under that context. Here we take the same object and ask what happens as the encoder absorbs progressively more of the estimation pipeline.

This is desirable because data collection and model estimation are often far more costly than model inference. Plotting the progression of estimation methods, classical methods require a sufficient amount of data for every task. Transfer learning and meta-learning are formulated to achieve similar performance with fewer samples, but still require explicit gradient-based estimators. Pushing this to its limit produces amortized estimators, such as context encoders, which only use a task's context to infer the task-specific data distribution and produce in-context inferences (Figure 8).

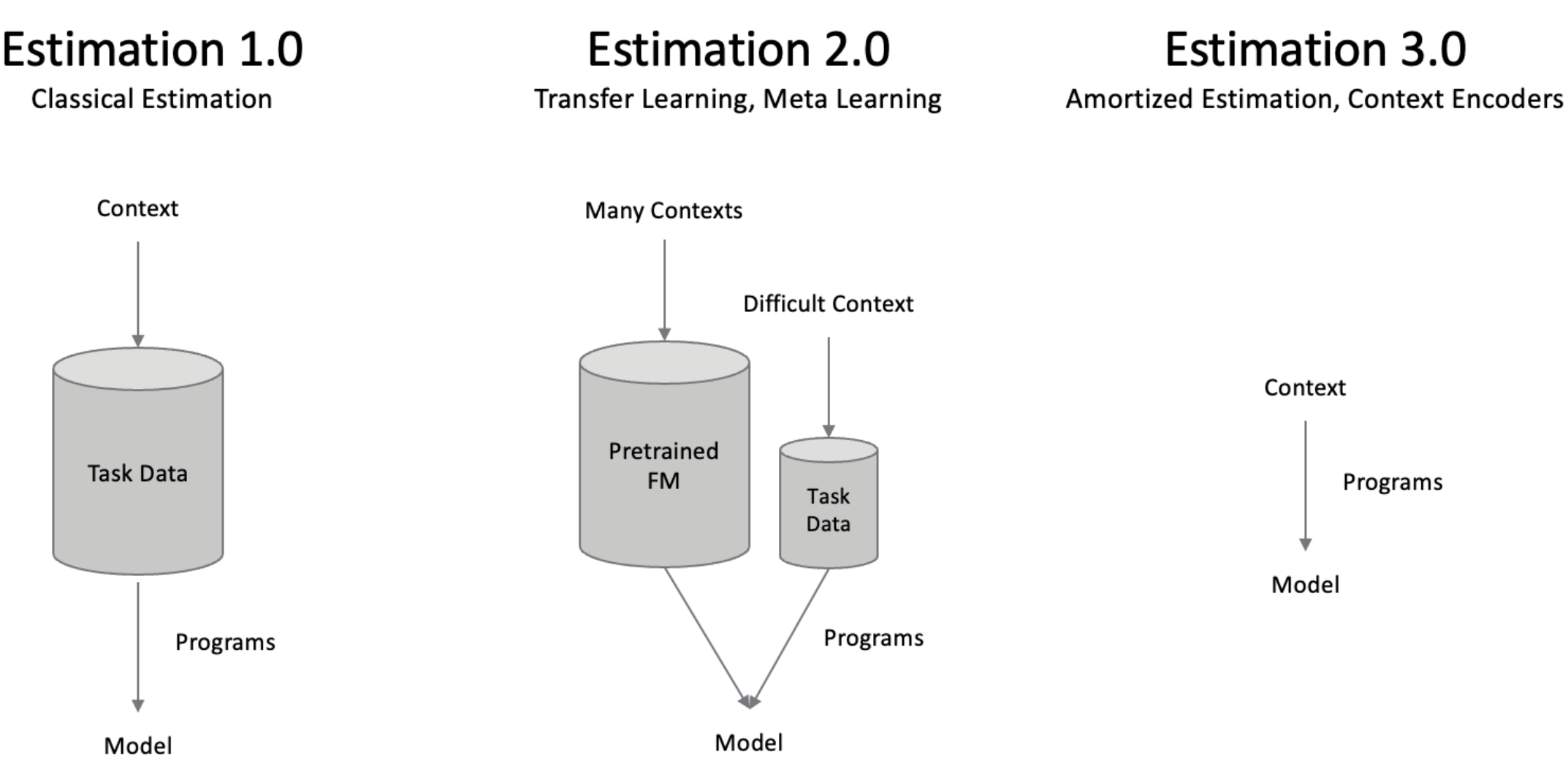

**Figure 8:** Evolution of Statistical Estimation. Classical estimators (left) require sufficiently large context-specific datasets for single-instance learning. Meta-learning and transfer learning (middle) decrease the need for context-specific data collection, but still perform a few rounds of gradient descent on context-specific data. Finally, amortized estimators (right) only use a representation of the data context to produce new model parameters, handling data collection and classical estimation implicitly.

In this regime, formal estimators are no longer required after the context encoder is obtained. We assume the context encoder can internally sample the data distribution and perform the necessary estimation steps entirely within a forward pass, similar to how amortized inference infers trajectories for expensive inference procedures. The key question becomes a practical one: When we don't need data, how do we encode the context of our hypothetical data distribution? Ironically, context itself is often hard to parameterize. A clearer representation of data distribution may be a few representative samples from the distribution itself. This leads to the practical implementation of implicit in-context learning with LLMs, where a few samples provided as natural language serve as context for implicit amortized estimation.

## In-Context Learning in Foundation Models

The most powerful and, arguably, most enigmatic form of implicit adaptivity is in-context learning (ICL), an emergent capability of large-scale foundation models. This phenomenon has become a central focus of modern AI research, as it represents a significant shift in how models learn and adapt to new tasks. This section provides an expanded review of ICL, beginning with a description of the core phenomenon, then deconstructing the key factors that influence its performance, reviewing the leading hypotheses for its underlying mechanisms, and concluding with its current limitations and open questions.

## The Phenomenon of Few-Shot In-Context Learning

First systematically demonstrated in large language models such as GPT-3 [116], ICL is the ability of a model to perform a new task after being conditioned on just a few examples provided in its input prompt. Critically, this adaptation occurs entirely within a single forward pass, without any updates to the model's weights. For instance, a model can be prompted with a few English-to-French translation pairs and then successfully translate a new word, effectively learning the task on the fly. This capability supports a broad range of applications, including few-shot classification, following complex instructions, and even inducing and applying simple algorithms from examples. Subsequent work has shown that the ability to generalize from few in-context examples can itself be enhanced through meta-training. MetaICL explicitly trains models across diverse meta-tasks, teaching them to infer and adapt within context at test time without gradient updates, thereby strengthening the implicit adaptability of large language models [117]. TabICL introduces a foundation model architecture for large-scale tabular data, showing that in-context learning can efficiently scale to structured datasets via column-row attention mechanisms [118]. These results suggest that implicit adaptivity generalizes beyond text or vision into the broader landscape of structured scientific data.

## Deconstructing ICL: Key Influencing Factors

The effectiveness of ICL is not guaranteed and depends heavily on several interacting factors, which have been the subject of extensive empirical investigation.

**The Role of Scale.** A critical finding is that ICL is an emergent ability that appears only after a model surpasses a certain threshold in scale (in terms of parameters, data, and computation). Recent work has shown that larger models do not just improve quantitatively at ICL; they may also learn in qualitatively different ways, suggesting that scale enables a fundamental shift in capability rather than a simple performance boost [119].

**Prompt Engineering and Example Selection.** The performance of ICL is highly sensitive to the composition of the prompt. The format, order, and selection of the in-context examples can dramatically affect the model's output. Counterintuitively, research has shown that the distribution of the input examples, rather than the correctness of their labels, often matters more for effective ICL. This suggests that the model is primarily learning a task format or an input-output mapping from the provided examples, rather than learning the underlying concepts from the labels themselves [120].

## Hypothesized Mechanisms: How Does ICL Work?

The underlying mechanisms that enable ICL are not fully understood and remain an active area of research. Several leading hypotheses have emerged, viewing ICL through the lenses of meta-learning, Bayesian inference, and specific architectural components.

**ICL as Implicit Meta-Learning.** The most prominent theory posits that transformers learn to implement general-purpose learning algorithms within their forward pass. During pre-training on vast and diverse datasets, the model is exposed to a multitude of tasks and patterns. This process is thought to implicitly train the model as a meta-learner, allowing it to recognize abstract task structures within a prompt and then execute a learned optimization process on the provided examples to solve the task for a new query [121,122].

**ICL as Implicit Bayesian Inference.** A complementary and powerful perspective understands ICL as a form of implicit Bayesian inference. In this view, the model learns a broad prior over a large class of functions during its pre-training phase. The in-context examples provided in the prompt act as evidence, which the model uses to perform a Bayesian update, resulting in a posterior predictive distribution for the final query. This framework provides a compelling explanation for how models can generalize from very few examples [123]. A complementary theoretical development interprets in-context learning as a rational adaptation process. From a Bayesian decision-theoretic standpoint, transformers can be viewed as implicitly balancing expected loss with strategy complexity, thereby achieving near-optimal adaptation under computational constraints [124]. This rational framing connects implicit adaptivity with classical principles of statistical inference.

**The Role of Induction Heads.** From a more mechanistic, architectural perspective, researchers have identified specific attention head patterns, dubbed “induction heads,” that appear to be crucial for ICL. These specialized heads are hypothesized to form circuits that can scan the context for repeated patterns and then copy or complete them, providing a basic mechanism for pattern completion and generalization from in-context examples [125]. Extending this mechanistic line, Dherin et al. (2025) demonstrate that stacking self-attention and MLP layers allows transformers to implicitly update internal representations during a single forward pass, effectively realizing dynamic context-specific weight adjustments without explicit training [126]. Such implicit internal updates offer a concrete mechanistic account of how context-dependent behavior arises.

### Limitations and Open Questions

Despite its remarkable capabilities, ICL faces significant limitations with respect to transparency, explicit control, and robustness. The adaptation process is opaque, making it difficult to debug or predict failure modes. Furthermore, performance can be brittle and highly sensitive to small changes in the prompt. As summarized in recent surveys, key open questions include developing a more complete theoretical understanding of ICL, improving its reliability, and establishing methods for controlling its behavior in high-stakes applications [127].

## Comparative Synthesis: Implicit versus Explicit Adaptivity

Implicit and explicit strategies reflect two complementary philosophies for modeling heterogeneity, each with distinct strengths and trade-offs. The optimal choice between these approaches depends on the goals of analysis, the structure and scale of available data, and the need for interpretability or regulatory compliance in the application domain.

**Implicit Adaptivity.** The principal advantage of implicit methods lies in their remarkable flexibility and scalability. Leveraging large-scale pre-training on diverse datasets, these models can effectively adapt to high-dimensional and unstructured contexts, such as raw text, images, or other complex sensory data, where explicitly specifying a context function $f(c)$ is infeasible. Because adaptation is performed internally during the model’s forward pass, inference is both rapid and adaptable. However, the mechanisms underlying this adaptability are typically opaque, making it challenging to interpret or control the model’s decision process. In applications like healthcare or autonomous systems, this lack of transparency can hinder trust, validation, and responsible deployment.

**Explicit Adaptivity.** In contrast, explicit models provide direct, interpretable mappings from context to parameters through functions such as $f(c)$. This structure supports clear visualization, statistical analysis, and the formulation of scientific hypotheses. It also enables more direct scrutiny and control of the model's reasoning. Nevertheless, explicit methods rely heavily on domain expertise to specify an appropriate functional form, and may struggle to accommodate unstructured or highly complex context spaces. If the assumed structure is misspecified, the model's performance and generalizability can be severely limited.

In summary, these two paradigms illustrate a fundamental trade-off between expressive capacity and transparent reasoning. Practitioners should carefully weigh these considerations, often choosing or blending approaches based on the unique demands of the task. For clarity, a comparative table or figure can further highlight the strengths and limitations of each strategy across various real-world applications.

## Open Challenges and the Motivation for Interpretability

The rise of powerful implicit adaptation methods, particularly in-context learning, raises critical open research questions regarding their diagnosis, control, and reliability. As these models are deployed in increasingly high-stakes applications, understanding their failure modes is not just an academic exercise but a practical necessity [60]. It is important to develop systematic methods for assessing when and why in-context learning is likely to fail, and to create techniques for interpreting and, where possible, steering the adaptation process. Prompting strategies such as chain-of-thought demonstrate that structured context can sometimes steer internal computation, providing limited but useful handles on model behavior [128]. A thorough understanding of the theoretical limits and practical capabilities of implicit adaptivity remains a central topic for ongoing research.

These considerations motivate a growing search for techniques that can make the adaptation process more transparent by "making the implicit explicit." Such methods aim to bridge the gap between the powerful but opaque capabilities of implicit models and the need for trustworthy, reliable AI. This research can be broadly categorized into several areas, including post-hoc interpretability approaches that seek to explain individual predictions [129], surrogate modeling where a simpler, interpretable model is trained to mimic the complex model's behavior, and strategies for extracting modular structure from trained models. A prime example of the latter is the line of work probing language models to determine if they have learned factual knowledge in a structured, accessible way [130]. By surfacing the latent structure inside these systems, researchers can enhance trust, promote modularity, and improve the readiness of adaptive models for deployment in real-world settings. This line of work provides a conceptual transition to subsequent sections, which explore the integration of interpretability with adaptive modeling.

# 5. Theoretical Bridges Between Varying-Coefficient Models and In-Context Learning

Recent theoretical work suggests that "explicit" context models (e.g., varying-coefficients, hierarchical/multitask) and "implicit" mechanisms (e.g., in-context learning via attention) often implement the same estimator class under squared loss, differing mainly in how they encode neighborhoods and regularization. Although these approaches arise from different traditions — one grounded in semi-parametric statistics, the other in large-scale deep learning — they implement strikingly similar estimators. This section formalizes these parallels and reviews key theoretical results establishing these bridges. #### Varying-Coefficient Models as Kernel Regression

Consider a semi-parametric varying-coefficient model in which each observation is governed by a parameter vector $\theta_i$ that depends smoothly on context $c_i$. For a new query context $c^*$, the parameter estimate is obtained by solving a locally weighted likelihood problem:

$$\widehat{\theta}(c^*) = \arg\max_{\theta} \sum_{i=1}^{n} K_\lambda(c_i, c^*)\, \ell(x_i; \theta),$$

where $K_\lambda$ is a kernel function that measures similarity between contexts and $\ell$ is the log-likelihood.

For regression with squared loss, this reduces to kernel ridge regression in the context space. Let $y = (y_1, \ldots, y_n)^\top$ and $K \in \mathbb{R}^{n \times n}$ be the Gram matrix with $K_{ij} = k(c_i, c_j)$. The prediction at $c^*$ is

$$\widehat{y}(c^*) = k(c^*)^\top (K + \lambda I)^{-1} y,$$

where $k(c^*) = (k(c^*, c_1), \ldots, k(c^*, c_n))^\top$. This expression highlights that varying-coefficient models perform kernel smoothing in the context space: nearby observations in context have greater influence on the parameter estimates at $c^*$.

Equivalently, the fitted model can be written as

$$\widehat{f}(x^*, c^*) = \sum_{i=1}^{n} \alpha_i(c^*)\, y_i,$$

where $\alpha_i(c^*)$ are normalized kernel weights determined entirely by the context similarities and the regularization parameter $\lambda$. This context-space kernel smoother is a special case of Proposition 1 from the introduction in which the input-space kernel is held fixed, so only context similarity is allowed to vary; the full proof is given in Appendix A.

## Transformers as Ridge and Kernel Regressors In-Context

A parallel line of research has demonstrated that transformers trained on simple regression tasks can learn to perform ridge or kernel regression entirely within their forward pass, without any explicit supervision to do so.

Akyürek et al. (2022) show that for linear regression tasks, transformers can learn to implement the ridge regression estimator

$$\widehat{w} = (X^\top X + \lambda I)^{-1} X^\top y$$

directly from a sequence of in-context examples. Each example $(x_i, y_i)$ is represented as a token, and the query token attends to the support tokens to compute the prediction for $x^*$; the attention mechanism learns to encode the solution to the regression problem [131].

Building on this finding, von Oswald et al. (2023) show that gradient-based training of transformers over distributions of regression tasks leads them to perform in-context gradient descent, effectively realizing kernel regression with the learned attention kernel serving as $k(c_i, c_j)$ [132]. Garg et al. (2023) further analyze which function classes can be learned in-context, demonstrating that transformers can approximate a wide family of kernel smoothers when trained on synthetic regression tasks [133].

Dai et al. (2023) provide a complementary theoretical view, arguing that transformers can implicitly implement compositional function families through their attention layers, and that in-context learning arises naturally from this functional representation [121].

Finally, Reuter et al. (2025) propose a compelling Bayesian interpretation: transformers trained under in-context learning can perform full Bayesian inference for common statistical models such as generalized linear models and latent factor models. Concretely, they train transformers to infer complex posterior distributions in context, showing that the in-context forward pass can approximate posterior sampling comparable to MCMC or variational inference methods [134].

In all these cases, the support set within the prompt plays an analogous role to the neighborhood in context space in varying-coefficient models. The query token's prediction is formed by aggregating information from the support tokens via learned similarity weights, realized by the attention mechanism rather than an explicitly defined kernel function.

### Two Paths to the Same Estimators

Taken together, these results reveal a common form:

$$\widehat{f}(x^*, c^*) = \sum_{i=1}^{n} \alpha_i(c^*)\, y_i,$$

where the weights $\alpha_i(c^*)$ depend on the relationship between the query context $c^*$ and support contexts $\{c_i\}$.

- In varying-coefficient models, $\alpha_i(c^*)$ are determined explicitly by a user-chosen kernel $K_\lambda$.
- In transformers, $\alpha_i(c^*)$ emerge implicitly from the learned attention patterns and internal computations after pretraining.

Both perspectives yield estimators of the same functional form, with explicit kernel weighting in VCMs and learned attention weighting in transformers. This correspondence motivates a unified view of context-adaptive inference, combining the interpretability of explicit modeling with the flexibility and scale of implicit computation.

## Theoretical Bridge

In practice, explicit methods provide interpretability and structure, while implicit methods provide flexibility and scalability. Understanding how explicit and implicit methods correspond offers a promising path toward adaptive, interpretable models at scale. Here, we focus on developing a generalized functional form that covers both explicit and implicit adaptation modes with a deep theoretical connection.

We study supervised prediction with units $i = 1, \ldots, n$. Each unit has a context $c_i \in \mathcal{C}$ (e.g., patient/user/site/time) and observed data $\mathcal{D}_i = \{(x_{ij}, y_{ij})\}_{j=1}^{m_i}$ with $x_{ij} \in \mathcal{X}$ and $y_{ij} \in \mathcal{Y}$. Predictions come from a model family $\mathcal{H} = \{h_\theta : \mathcal{X} \to \mathcal{Y} \mid \theta \in \Theta\}$ (e.g., linear model, neural net, probabilistic model).

In global (i.i.d.) models, $\theta_i \equiv \theta^\star$. In context-adaptive models, parameters vary with context: $\theta_i = f(c_i)$ or $\theta_i \sim P(\theta \mid c_i)$.

For a new unit with context $c$, we write a unified empirical objective:

$$\widehat{\theta}(c) \in \arg\min_{\theta\in\Theta} \underbrace{\sum_{(i,j)\in S(c)} \ell\big(h_\theta(x_{ij}), y_{ij}\big)}_{\text{context-dependent support}} + \underbrace{\mathcal{R}(\theta;\, c)}_{\text{context-structured regularization}},$$

(1) where $\ell$ is a proper loss (e.g., squared, logistic), $S(c) \subseteq \{1, \ldots, n\} \times \mathbb{N}$ is a support set selected for context $c$, and $\mathcal{R}(\theta; c)$ encodes how parameters are allowed to vary with context (smoothness, sparsity, low-rank, hierarchy, etc.). This decomposition pairs a context-weighted loss, as in local likelihood and kernel-weighted estimation [1], with a context-structured regularizer from the penalized-estimation literature. This form has been applied to structured, high-dimensional context-varying estimation and comes with sparsistency guarantees that make it a central tool for context-adaptive modeling [6,64].

This form provides two ways to inject contextual information: - **Explicit parameterization:** a map $f : \mathcal{C} \to \Theta$ sets $\theta_i = f(c_i)$ (e.g., varying-coefficients, hierarchical Bayes, multi-task/meta-learning). Here $\mathcal{R}(\theta; c)$ typically regularizes $f$ (e.g., Lipschitz over $\mathcal{C}$, group lasso, low-rank). - **Implicit parameterization:** context alters optimization or internal states without exposing $\theta$ directly (e.g., mixture-of-experts with gates $g(x, c)$; retrieval where $S(c)$ is built by a retriever; in-context learning where a prompt map $S(c)$ conditions a foundation model).

Emerging approaches blur this distinction. Instead of treating prompting (implicit) and fine-tuning (explicit) as separate mechanisms, models can map contextual information, such as a natural language task description, directly into parameter updates. Text-to-LoRA provides a concrete example of this paradigm, where context is encoded and used to generate task-specific parameters, effectively collapsing the boundary between implicit and explicit adaptation [135].

To formalize this connection, we now return to a unified estimator. For convenience, we use a context encoder $\phi : \mathcal{C} \to \mathbb{R}^d$ and a similarity/kernel $K(c, c')$. A common instance of Eq. 1 is kernel-weighted risk:

$$\sum_{i,j} w_{ij}(c)\, \ell\big(h_\theta(x_{ij}), y_{ij}\big) + \mathcal{R}(\theta), \qquad w_{ij}(c) \propto K\big(\phi(c), \phi(c_i)\big) \cdot \mathbf{1}\big[(i,j) \in S(c)\big].$$

Adaptivity across estimation and inference is governed by three tools that can be applied independently: 1) **Information** via $S(c)$ (what context is exposed),
2) **Inductive bias** via $\mathcal{R}(\theta; c)$ (how parameters may vary),
3) **Compute** via warm-starts/caching/steps (how aggressively we solve Eq. 1 at test time).

## Proof Overview

To formalize this link between explicit and implicit context adaptation, we require a few assumptions.

1. Exchangeability within context: conditional on $(\theta_i, c_i)$, $(x_{ij}, y_{ij})$ are i.i.d.
2. Regularity: either $\theta = f(c)$ with $f$ in a regular class (e.g., Lipschitz/sparse/low-rank) or retrieval weights $w_{ij}(c)$ are bounded and locally normalized.
3. Identifiability/stability: $\ell$ is convex in model outputs and $\mathcal{R}$ yields a unique or stable minimizer.
4. Resource tracking: we track $|S(c)|$, optimization steps, and memory to compare **adaptation efficiency**.

**Proposition 1 (Explicit varying-coefficients and linear ICL coincide with kernel ridge on joint features in the linear squared-loss setting).**
Assume squared loss and the regression model $y = \langle \theta(c), x \rangle + \varepsilon$ with $\mathbb{E}[\varepsilon] = 0$. Let (i) a context encoder $\phi : \mathcal{C} \to \mathbb{R}^{d_c}$, (ii) joint features $\psi(x, c) := x \otimes \phi(c) \in \mathbb{R}^{d_x d_c}$, where $\otimes$ is the Kronecker product, and $d_x$ and $d_c$ are the dimensions of $x$ and $\phi(c)$, (iii) a context-dependent support set $S(c)$ with nonnegative weights $w_{ij}(c)$.

- **(A) Explicit varying-coefficients.** Let $\theta(c) = B\,\phi(c)$ with $B \in \mathbb{R}^{d_x \times d_c}$ and ridge penalty $\lambda \|B\|_F^2$. The weighted ridge solution yields

$$\widehat{y}(x, c) = k_{(x,c)}^{\top}\big(K + \lambda I\big)^{-1} y, \quad K_{ab} = \langle \psi_a, \psi_b \rangle = \langle x_a, x_b \rangle \cdot \langle \phi(c_a), \phi(c_b) \rangle,$$

  i.e., **kernel ridge regression (KRR)** on joint features.

- **(B) Implicit adaptation via linear ICL.** Let a single linear attention layer consume the weighted support set $S(c)$ with linear query, key, and value maps $q = Q\psi$, $k = K\psi$, $v = V\psi$ (projection matrices $Q$, $K$, $V$, distinct from the Gram matrix $K$ above) and a linear readout. With attention weights proportional to $w_{ij}(c) \cdot \langle q, k_{ij} \rangle$, the induced predictor equals KRR with kernel

$$k\big((x, c), (x', c')\big) = \langle q(x, c), k(x', c') \rangle,$$

  i.e., a learned **dot-product kernel** on the same joint features. If attention parameters are trained in the linearized/NTK regime, learning equals kernel regression with the network's NTK, which is again a dot-product kernel on linear transforms of $\psi$.

**Corollary 1 (Retrieval, gating, and weighting are kernel/measure choices).** Choosing $S(c)$ via a retriever $R(c)$, or gating in a mixture-of-experts, corresponds to changing the kernel and/or the empirical measure (weights $w_{ij}(c)$) used by KRR on $\psi$.

Full proof in Appendix A.

## Positioning and prior art

Proposition 1 is expository: part (A) is standard ridge $\leftrightarrow$ kernel duality on joint features; part (B) follows from (i) fixed attention + trained linear head = ridge on fixed features and (ii) NTK linearization $\to$ kernel regression with the network's NTK. Our contribution is the unified context-aware formulation: explicit design tools via $S(c)$, $\mathcal{R}(\theta; c)$, and compute, and the mapping of retrieval/gating to kernel/measure choices. See transformer ICL as classical estimators [136,137,138] and NTK analyses [139,140].

## Limitations

This linear, squared-loss bridge captures a large class of explicit and implicit adaptors, but it abstracts away at least three realities: (i) non-quadratic losses (e.g., logistic) change the effective kernel/weighting via loss curvature; (ii) when the prediction head is nonlinear (e.g., an MLP or attention), it introduces model curvature: the head's Jacobian/Hessian depends on the input and parameters, so the effective metric and weights change with the representation; the fixed-kernel view from the linear case no longer holds; (iii) multi-modal context encoders (text, graphs, images) alter both the neighborhood definition and the regularizer. We view these as open extensions to explore beyond the scope of this review.

# 6. Principles of Context-Adaptive Inference

What makes a model adaptive? When is it good for a model to be adaptive? While the appeal of adaptivity lies in flexibility and personalized inference, not all adaptivity is beneficial. This section formalizes the core principles that underlie adaptive modeling and situates them within both classical statistics and recent advances in machine learning.

Adaptivity is best understood as a structured set of design principles rather than a single mechanism, each a different axis along which a model can incorporate or restrict adaptation. We organize the section around six: flexibility, heterogeneity signals, modularity, selectivity, data efficiency, and tradeoffs.

These principles are realized by the two main method families in different ways. Explicit methods specify the map $\theta = f(c)$ directly and estimate it with structured regularizers; implicit methods let the same adaptation emerge inside a large model's forward pass. The principles we propose are grounded in Eq. 1 so they are generally applicable.

## 1. Adaptivity requires flexibility

The first principle concerns model capacity. A model must be able to represent multiple behaviors if it is to adapt. Without sufficient representational richness, adaptation becomes superficial, amounting only to noise-fitting rather than meaningful personalization. Flexibility provides the foundation for models to express diverse responses across individuals, groups, or environments, rather than enforcing a single global rule.

Flexibility may arise from different modeling strategies. In classical statistics, regression models with interaction effects explicitly capture how predictors influence outcomes differently across contexts, while hierarchical and multilevel models let effects vary systematically across groups. Varying-coefficient models extend this further by allowing regression coefficients to evolve smoothly with contextual covariates [3]. In machine learning, meta-learning and mixture-of-experts architectures [141] offer dynamic allocation of capacity, training models to specialize on tasks or inputs as needed. Without flexibility, adaptation has no space in which to operate. In Eq. 1, flexibility is the expressiveness of the model family $\{h_\theta\}$ and the room left for $\theta(c)$ to move; the explicit and implicit sections trace two routes to it, structured maps $f(c)$ and large architectures that discover it emergently through internal weights.

## 2. Adaptivity requires a signal of heterogeneity

Flexibility alone is not enough; a model also requires observable signals that indicate how and why adaptation should occur. Without such signals, adaptive systems risk reacting to random fluctuations rather than capturing meaningful structure. In statistics, varying-coefficient regressions illustrate this idea by allowing parameters to change smoothly with observed covariates [3], while hierarchical models assume systematic group differences that provide a natural signal for adaptive pooling.

In machine learning, contextual bandits adapt decisions to side information that characterizes the current environment, while benchmarks like WILDS highlight that real-world datasets often contain distributional shifts and subgroup heterogeneity [142]. Recent work extends this further, modeling time-varying changes in continuous temporal domain generalization [143] or using diversity across experts to separate stable from unstable patterns [144]. Across applications, from medicine to online platforms, signals that indicate heterogeneity or distribution shift drive adaptation.

Causal inference is a canonical example. In the Neyman-Rubin potential-outcomes framework, the average treatment effect $E[Y(1) - Y(0)]$, the mean difference between the potential outcomes under treatment $Y(1)$ and control $Y(0)$, assumes one effect shared across the population. If the underlying population contains multiple groups (heterogeneous) and the effect differs between groups, the average estimator is confounded. Recovering heterogeneous treatment effects across groups requires side information $C$ about group membership, i.e. “no unmeasured confounding” (Figure 9).

Average Treatment Effect

$E[Y(1) - Y(0) \mid X]$

Conditional Average Treatment Effect

$E[Y(1) - Y(0) \mid X, C]$

**Figure 9:** Heterogeneous treatment effects. Left: average treatment effect (ATE) conditional on $X$, implicitly assuming homogeneity across contexts. Right: conditional average treatment effect (CATE) that allows treatment effects to vary systematically with additional context $C$.

In Eq. 1, this principle is encoded by $S(c)$: a signal of heterogeneity is what makes the support set localized rather than an arbitrary neighborhood.

## 3. Modularity improves adaptivity

Organizing adaptation into modular units improves interpretability and robustness. Instead of spreading changes across an entire system, modularity restricts variation to well-defined subcomponents that can be recombined, reused, or replaced. This structure provides three advantages: targeted adaptation, transferability across tasks, and disentanglement of variation sources.

A canonical example is the mixture-of-experts framework, where a gating network routes inputs to specialized experts trained for different data regimes [141]. By decomposing capacity in this way, models not only gain efficiency but also clarify which components are responsible for specific adaptive behaviors. Recent advances extend this principle in modern architectures: modular domain experts [145], adapter libraries for large language models [146], and mixtures of LoRA experts [147]. In applications from language processing to computer vision, modularity is now central to scaling adaptation.

In Eq. 1, $\mathcal{R}(\theta; c)$ structures on how $\theta$ may vary, e.g. through block, low-rank, or n-experts constraints $\mathcal{R}(\theta; c)$.

## 4. Adaptivity implies selectivity

Adaptation must not occur indiscriminately. Overreacting to noise leads to overfitting, defeating the purpose of adaptation. Selectivity provides the discipline that ensures adaptive mechanisms respond only when supported by reliable evidence.

Classical statistics formalized this principle through methods such as Lepski's rule for bandwidth selection, which balances bias and variance in nonparametric estimation [148]. Aggregation methods such as the weighted majority algorithm show how selective weighting of multiple models can improve robustness [149]. In modern machine learning, Bayesian rules can activate test-time updates only when uncertainty is manageable [150], while confidence-based strategies prevent unstable adjustments by holding back adaptation under weak signals [151]. Sparse expert models apply the same principle architecturally, activating only a few experts for easy inputs but engaging more capacity for difficult cases [152]. These safeguards demonstrate that good adaptation is selective adaptation.

in Eq. 1, the regularizer $\mathcal{R}(\theta; c)$ may also act as a brake, deciding when the support set is allowed to pull $\theta(c)$ away from a shared value, whether through a bandwidth, a confidence threshold, or a sparse gate.

## 5. Adaptivity is bounded by data efficiency and data volume

Even with flexibility, heterogeneity, modularity, and selectivity in place, the scope of adaptation is fundamentally constrained by the amount of relevant data. Fine-grained adaptation requires sufficient samples to estimate context-specific effects reliably. When data are scarce, adaptive systems risk inflating variance, capturing noise, or overfitting. This basic statistical constraint is the fundamental bias-variance tradeoff [153], but it occurs more acutely in context-adaptive models due to (i) the expectation that information will transfer from well-samples contexts to under-sampled contexts and (ii) the ability of context-adaptive models to sample much larger and more heterogeneous data spaces from loosely connected tasks.

The traditional way of addressing this is Eq. 1's regularization term $\mathcal{R}(\theta; c)$: increasing regularization lowers variance, compute, and instability, but raises bias. Unlike traditional models, context-adaptive models can also address this by expanding datasets to train across more diverse contexts, increasing the effective size of $S(c)$ to combat variance (Figure 10).

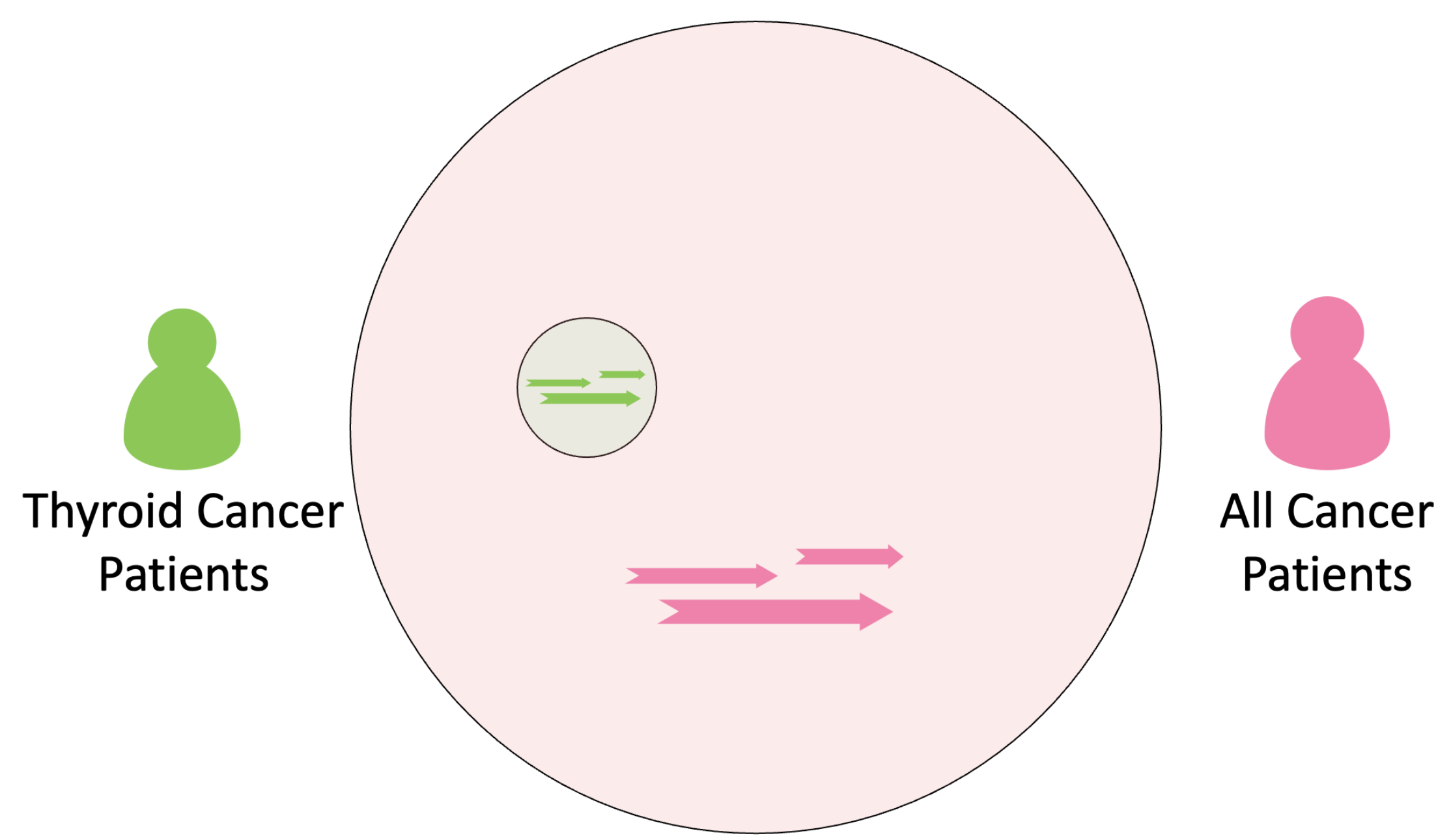

**Figure 10:** Context-adaptive models use context signals to understand specific subpopulations or tasks (e.g. modeling thyroid cancer patients) within broader, heterogeneous populations or task sets (e.g. all cancer patients) during training. At inference time, context localizes the model to a specific predictive task. Context-adaptive models have higher

variance than task-specific models, but context allows the model to ingest larger and more heterogeneous datasets, offsetting this cost.

A recent analysis on varying-coefficient models explores scaling laws for context-adaptive models in terms of two axes: “vertical scaling,” the traditional mode where model performance improves with task-specific samples, and “horizontal scaling,” where model performance improves by incorporating samples from new tasks [97]. This analysis found that (i) under a fixed sample budget, horizontal scaling can outperform vertical scaling (ii) this happens at a critical point where the model has seen sufficiently many tasks, and begins to generalize to new tasks.

Meta-learning research also illustrates this tension, as few-shot frameworks show both the promise of cross-task generalization and the sharp degradation that occurs when task diversity or sample size is insufficient [154]. Bayesian analyses of scaling laws for in-context learning formalize how the reliability of adaptation grows with data [155]. More generally, context-adaptive models present new opportunities for study design, allowing controlled task-specific data collection to be substituted for large heterogeneous data from diverse sources.

## When Adaptivity Fails: Common Failure Modes

The six principles describe when adaptation should succeed, but in practice, failures remain common. Understanding these failure modes is important for designing safeguards, since they reveal how adaptive methods break when principles are ignored or misapplied. Failure does not imply that models lack adaptivity, but that adaptation proceeds in unstable or unjustified ways.

**Spurious adaptation.** Models sometimes adapt to unstable or confounded features that predict the outcome within the training environments but do not hold across them. This is closely related to shortcut learning in deep networks, where spurious correlations masquerade as useful signals [142,156]. Crucially, this is not a symptom of mis-tuned complexity: the offending correlation can be strong, low-variance, and stable under more data, and a model with more or fewer parameters latches onto the same shortcut. What fails is the direction of adaptation, not its degree. In terms of Eq. 1, when the regularizer $\mathcal{R}(\theta; c)$ does not constrain adaptation toward invariant structure, the estimator conditions on whichever context feature most reduces in-distribution loss, so a model that is over-eager to adapt by design will prefer a strong spurious signal to a weaker stable one. The failure surfaces only when the environment shifts and the correlation reverses or disappears. Unlike overfitting, collecting more data from the same distribution does not remove it; only a change of training environments or an explicit invariance constraint does, the remedy taken up in the context-invariant section.

**Overfitting in low-data contexts.** Fine-grained adaptation requires sufficient signal. When the available data are limited, adaptive models inflate variance and personalize to noise rather than meaningful structure. Meta-learning research illustrates this tension: although few-shot methods aim to generalize with minimal samples, they often degrade sharply when task diversity is low or heterogeneity is weak [154]. Unlike spurious adaptation, this failure is related to capacity: the underlying structure is real but the support $S(c)$ is too small to resolve it, so more data, stronger regularization, or coarser contexts are needed. This underscores how data efficiency sets unavoidable limits on adaptivity.

**Modularity mis-specification.** Although modularity can improve interpretability and transfer, poorly designed modules or unstable routing mechanisms can create new sources of error. Group-shift robustness studies reveal that when partitions are misaligned with true structure, adaptive pooling can worsen disparities across groups [157]. Similarly, analyses of mixture-of-experts models show that mis-specified routing can cause experts to collapse or remain underutilized [158]. These cases highlight that modularity is beneficial only when aligned with meaningful heterogeneity.

**Feedback loops.** Adaptive models can also alter the very distributions they rely on, especially in high-stakes applications such as recommendation, hiring, or credit scoring. This creates feedback loops where bias is reinforced rather than corrected. For example, an adaptive recommender system that over-personalizes may restrict exposure to diverse content, reshaping user behavior in ways that amplify initial bias. The selective labels problem in algorithmic evaluation illustrates how unobserved counterfactuals complicate learning from adaptively collected data [159]. These examples show that adaptation must be evaluated with attention to long-term interactions, not only short-term accuracy.

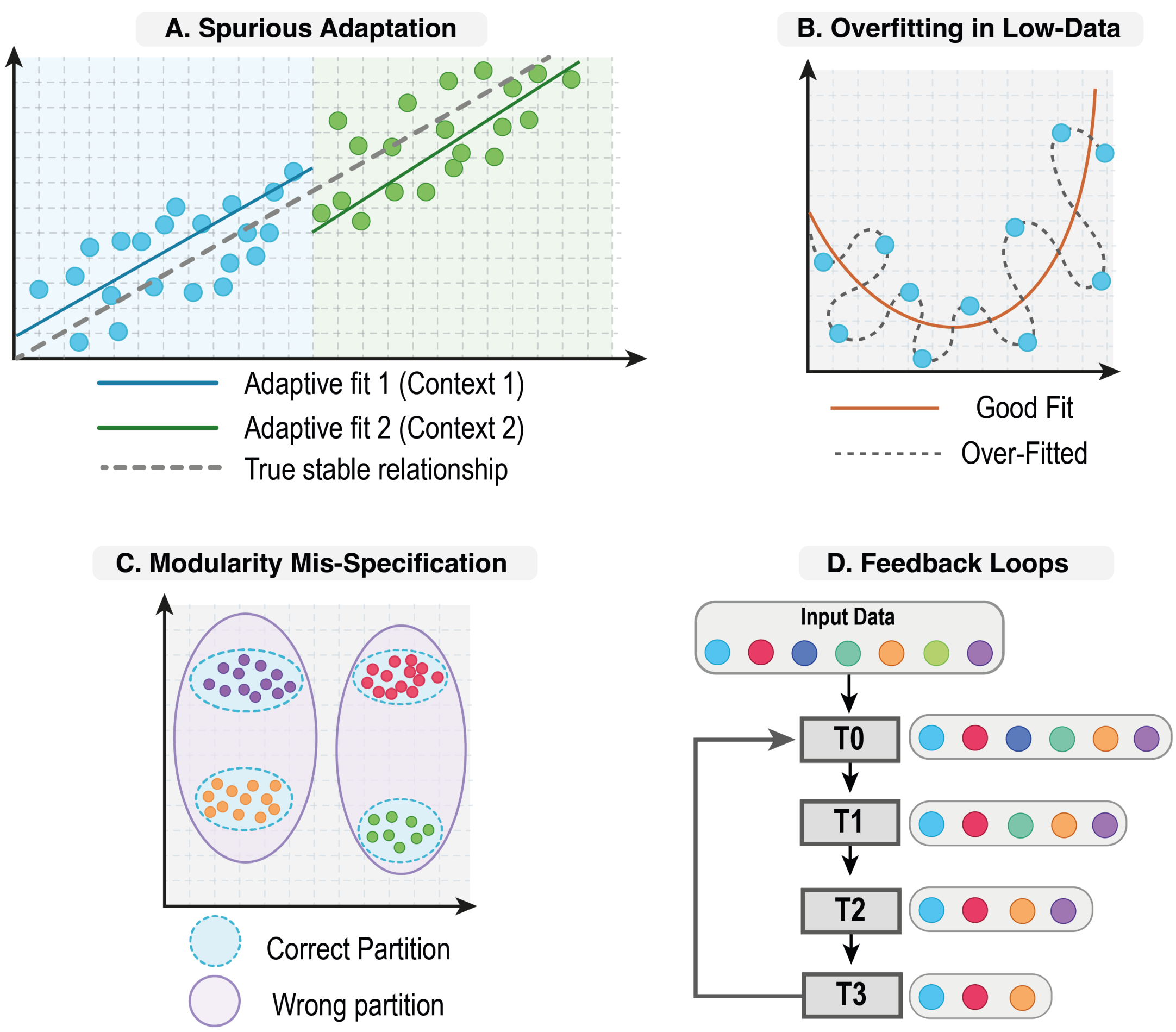

**Figure 11:** Failure Modes of Context-Adaptive Models. (A) **Spurious Adaptation**: the fit tracks a feature that predicts in the training environment but reverses under shift. (B) **Overfitting in Low-Data Contexts**: adaptation follows noise rather than signal. (C) **Modularity Mis-Specification**: incorrect partitions obscure the true structure. (D) **Feedback Loops**: adaptive decisions reshape the very data they rely on.

## Implications

For practitioners, these insights translate into a design recipe. Begin by ensuring sufficient flexibility, but constrain it through modular structures that make adaptation interpretable and transferable. Seek out reliable signals of heterogeneity that justify adaptation, and incorporate explicit mechanisms of selectivity to guard against noise. Respect the limits imposed by data efficiency, recognizing that fine-grained personalization requires sufficient statistical support. Always weigh the tradeoffs explicitly, balancing personalization against stability, efficiency against interpretability, and short-term gains against long-term robustness. Evaluation criteria should extend beyond predictive accuracy to include calibration, fairness across subgroups, stability under distributional shift, and resilience to feedback loops.

# 7. Evaluation of Context-Adaptive Models

The preceding sections described how context-adaptive models are built, explicitly through a structured $f(c)$ and implicitly through emergent computation in large models. Both paradigms raise the same three practical questions once a model is deployed: how cheaply it adapts, how stably it routes a context to the right behavior, and how well that behavior holds up when the context shifts. This section collects the design principles and proposes formal evaluation metrics that answer these questions.

## Measuring Context-Conditional Performance

We base every metric below on a single straightforward quantity: performance conditioned on context. Let $\mathcal{C}$ denote the context space and $\mathcal{D}_{\text{test}}$ a test distribution over $(x, y, c)$. For a predictor $\hat{f}$, define the context-conditional risk as

$$\mathcal{R}(\hat{f} \mid c) = \mathbb{E}\Big[\,\ell\big(\hat{f}(x,c), y\big)\,\Big|\, c\,\Big], \qquad \mathcal{R}(\hat{f}) = \mathbb{E}_{c\sim\mathcal{D}_{\text{test}}}\Big[\,\mathcal{R}(\hat{f} \mid c)\,\Big].$$

A single aggregate risk hides exactly the heterogeneity these models exist to capture, so a context-stratified evaluation reports $\mathcal{R}(\hat{f} \mid c)$ across predefined bins or via a smoothed estimate $\int \mathcal{R}(\hat{f} \mid c)\, \mathrm{d}\Pi(c)$ for a reference measure $\Pi$ that weights regions of the context space.

## Adaptation Efficiency

The efficiency of a context-adaptive method hinges on design choices that trade computational tractability against statistical accuracy, and the same choices govern whether a method scales to large data while remaining interpretable.

One central principle is the use of sparsity assumptions to limit the number of context-dependent parameters. This can be achieved through group sparsity, which encourages entire groups of parameters to be zero simultaneously [160], hierarchical regularization that applies different strengths of shrinkage to varying levels of context specificity [161], and adaptive thresholding that adjusts sparsity levels in accordance with context complexity.

Efficiency can also be enhanced through computational strategies that allocate resources adaptively. Early stopping terminates optimization for contexts where convergence occurs rapidly [162], while context-dependent sampling employs different sampling schemes across contexts [163]. Caching and warm-starting further accelerate optimization by reusing solutions from similar contexts, which is particularly effective when contexts vary smoothly [164]. Related context-dependent computation appears in adaptive batching, per-context learning rates, and multi-fidelity pipelines that allocate compute and precision by context complexity.

To make these gains measurable, let $S_k(c) = \{(x_j, y_j, c)\}_{j=1}^k$ denote $k$ examples available within context $c$, and define the adaptation efficiency as

$$\mathrm{AE}_k(c) = \mathcal{R}(\hat{f}_0 \mid c) - \mathcal{R}(\hat{f}_{S_k} \mid c), \qquad \mathrm{AE}_k = \mathbb{E}_c[\,\mathrm{AE}_k(c)\,],$$

where $\hat{f}_0$ is the non-adapted baseline and $\hat{f}_{S_k}$ the adapted predictor; the curve $k \mapsto \mathrm{AE}_k$ summarizes few-shot adaptation gains across context sizes. A closely related quantity measures how much of this efficiency carries between contexts. For a shared representation $\phi$ transferred from source to target contexts $\mathcal{C}_{\mathrm{src}} \to \mathcal{C}_{\mathrm{tgt}}$, the transfer gain is

$$\mathrm{TG}(\phi) = \mathcal{R}_{\mathcal{C}_{\mathrm{tgt}}}\big(\hat{f}_{\mathrm{scratch}}\big) - \mathcal{R}_{\mathcal{C}_{\mathrm{tgt}}}\big(\hat{f}_{\phi}\big),$$

the risk reduction from transferring $\phi$ rather than training from scratch, so that positive values indicate useful transfer, matching the sign convention of $\mathrm{AE}_k$. This is the same efficiency argument that motivates the amortized context encoders of the previous sections.

## Formalization: data limits on adaptation efficiency

These gains are ultimately bounded by how much data each context supplies. Let contexts lie in a metric space $(\mathcal{C}, d)$, let the per-context parameter be $\theta(c)$, and for an observation $(x, y, c)$ let $p_\theta(y \mid x, c)$ be a conditional model with loss $\ell(\theta; x, y, c)$. For a neighborhood $\mathcal{N}_\delta(c) = \{c' : d(c, c') \le \delta\}$, the effective sample size available to estimate $\theta(c)$ is

$$N_{\mathrm{eff}}(c,\delta) = \sum_{i=1}^{n} w_\delta(c_i, c), \qquad w_\delta(c_i, c) \propto K\!\left(\frac{d(c_i, c)}{\delta}\right), \qquad \sum_i w_\delta(c_i, c) = 1,$$

for a kernel $K$. A kernel-regularized estimator with smoothness penalty

$$\mathcal{R}(\theta) = \int \|\nabla_c \theta(c)\|^2 \,\mathrm{d}c$$

solves

$$\widehat{\theta} = \arg\min_{\theta \in \Theta} \frac{1}{n} \sum_{i=1}^{n} \ell(\theta; x_i, y_i, c_i) + \lambda\, \mathcal{R}(\theta).$$

Under local Lipschitzness in $c$ and an $L$-smooth, $\mu$-strongly convex risk in $\theta$, a standard bias–variance decomposition yields for each component $j$

$$\mathbb{E}\Big[\|\widehat{\theta}_j(c) - \theta_j(c)\|^2\Big] \lesssim \underbrace{\frac{\sigma^2}{N_{\mathrm{eff}}(c,\delta)}}_{\text{variance}} + \underbrace{\delta^{2\alpha}}_{\text{approx. bias}} + \underbrace{\lambda^2}_{\text{reg. bias}}, \qquad \alpha > 0.$$

where $\sigma^2$ is the noise variance, $L$ and $\mu$ the smoothness and strong-convexity constants of the risk, and $\alpha > 0$ the context-smoothness exponent governing the approximation bias. Finer locality (small $\delta$) sharpens resolution but shrinks $N_{\mathrm{eff}}$ and inflates variance; this is the quantitative form of the adaptation–data tradeoff, and amortized encoders raise $N_{\mathrm{eff}}$ by sharing $\theta(c) = f_\phi(c)$ across contexts. Compute enters through the same neighborhood: an early-stopped first-order method with step size $\eta$ and $T(c)$ context-dependent iterations satisfies

$$\mathcal{L}(\theta^{(T(c))}) - \mathcal{L}(\theta^\star) \leq (1 - \eta\mu)^{T(c)} \big(\mathcal{L}(\theta^{(0)}) - \mathcal{L}(\theta^\star)\big) + \frac{\eta L \sigma^2}{2\mu\, N_{\text{eff}}(c, \delta)},$$

where $\mathcal{L}$ is the population objective, $\theta^{(0)}$ the initialization, $\theta^\star$ its minimizer, and $\eta$ the step size. This links the compute budget $T(c)$ and data availability $N_{\text{eff}}(c, \delta)$ to the attainable excess risk at context $c$.

## Routing Stability

Both explicit and implicit models adapt by choosing which parameters or which experts govern the prediction. Explicit models make this decision overtly, through the partitions or context encoders that map context to model parameters. Implicit models make the same decision internally, through the gating of a mixture of experts or the attention weights of a transformer. In either case the decision is a function of context, and the stability of this function plays a key role in generalization: small perturbations of the context should not cause erratic changes in which behavior is selected, or predictions become unpredictable near routing boundaries.

Instability is easiest to see at the boundaries. A partition model can flip a sample between adjacent leaves under a negligible change in a splitting variable, and an in-context learner can change its answer under a reordering or reformatting of the same prompt examples, a sensitivity documented for in-context learning where the distribution and arrangement of examples can matter more than their content [120]. In sparse expert models routing instability can cause experts to collapse or remain underused, undermining reliability [158]. Reporting routing stability therefore means probing the model with perturbations that should be behavior-preserving, such as reordering exchangeable examples, jittering a continuous context near a learned split, or relabeling an equivalent measurement policy, and measuring how much the selected behavior and the resulting prediction move. Smoothly routed models degrade gracefully across these perturbations while sharply partitioned or brittle ones do not. The difference is a design choice as much as a diagnostic.

This admits a single measure across both discrete and continuous perturbations. Let $\rho(c)$ denote the routing decision the model exposes, whether gate weights, a leaf assignment, or the adapted parameters $\theta(c)$, and let $\Pi_c$ be a family of behavior-preserving perturbations of context $c$. The routing stability at $c$ is the expected movement of that decision under perturbation,

$$\mathrm{RSt}(c) = \mathbb{E}_{\pi \sim \Pi_c}\, d\big(\rho(c),\ \rho(\pi(c))\big),$$

with a supremum over $\Pi_c$ giving the worst-case variant. Discrete invariances such as reordering exchangeable examples enter through $\Pi_c$ directly; for continuous perturbations, normalizing by the context displacement $D(c, \pi(c))$ recovers a local Lipschitz constant of the routing map, which is the smoothness reading of stability for continuous contexts.

## Context-Specific Robustness

The final major evaluation type asks whether adaptation still helps when the context distribution at test time differs from training. Context shift takes many forms, including new subpopulations, drifting covariates, and unseen measurement policies. For explicit models it is easiest to make concrete when context is the pattern of observed measurements, since the deployment environment routinely presents measurement policies that were rare or absent during training; we use missingness-as-context as the running example (i.e. input dropout). Evaluation of missingness-as-context models should report mask-stratified metrics, including worst-group performance, following group-robust evaluation practice [142,157]. Robustness should be probed with mask-shift stress tests, training under one measurement policy and testing under another, to quantify degradation and the benefit of contextualization, as formalized in the Domain Adaptation under Missingness Shift (DAMS) setting [142,165]. When imputation is used, authors should assess imputation realism by holding out observed entries under realistic mask distributions and reporting MAE/RMSE and calibration for $p(x_{\text{missing}} \mid x_{\text{observed}})$ [166,167]. For causal or estimation applications, conduct ignorability sensitivity analyses, contrasting MAR-based results with pattern-mixture or selection-model analyses under plausible MNAR mechanisms [168,169]. Finally, include ablations that remove mask or indicator inputs, and, for trees, disable default-direction routing, to confirm that gains derive from modeling the mask signal rather than from artifacts [170,171]. Practical implementations of these ideas are widely available: GRU-D [172] and BRITS [173] provide mask- and time-aware sequence models, while GAIN [167] and VAEAC [166] offer open-source code for imputation under arbitrary masks. For tree ensembles, XGBoost supports sparsity-aware default-direction splits, making it straightforward to treat missing values as context without preprocessing [174].

Implicit models face the same issue in a different form. In-context learning adapts within a single forward pass, but that adaptation is opaque and can be brittle: performance is sensitive to small changes in the prompt, and reliability under distribution shift is hard to guarantee or to audit. Recent surveys frame the open questions as developing a more complete theoretical account of when in-context learning fails, improving its reliability, and establishing methods for controlling its behavior in high-stakes applications [127]. The evaluation logic carries over from the explicit case: stratify performance by the kind of context, stress-test under context shift, and ablate the adaptive signal to confirm that the model is using context as intended rather than exploiting an artifact. From this perspective, the mask-shift stress test for an explicit model and the prompt-perturbation test for an in-context learner are the same experiment applied to two realizations of $f(c)$.

The degradation these tests reveal can be summarized in a single score. Let $Q$ denote a family of admissible context shifts, for example $f$-divergence or Wasserstein balls over context marginals; the robustness score

$$\mathrm{RS}(\hat{f}; Q) = \sup_{\widetilde{\mathcal{D}} \in Q} \left[ \mathcal{R}_{\widetilde{\mathcal{D}}}(\hat{f}) - \mathcal{R}_{\mathcal{D}_{\text{test}}}(\hat{f}) \right]$$

reports the worst-case excess risk over that family, with higher values indicating greater sensitivity to contextual change.

# 8. Toward Explicit Modeling of Implicit Adaptivity: Local Models, Surrogates and Post Hoc Approximations

Building on the prior discussion of implicit adaptivity, this section examines methods that expose, approximate, or control those adaptive mechanisms.
Implicit adaptivity allows powerful models, including foundation models, to adjust behavior without explicitly representing a mapping from context to parameters [60]. This flexibility obscures the underlying mechanisms of adaptation, hindering modular reuse and systematic auditing. Making adaptivity explicit improves alignment with downstream goals, enables modular composition, and supports debugging and error attribution. It also fits the call for a more rigorous science of interpretability with defined objectives and evaluation criteria [175,176].
This chapter reviews practical approaches for surfacing structure, the assumptions they rely on, and how to evaluate their faithfulness and utility.

## From Implicit to Explicit Adaptivity

Implicit adaptivity is hidden, flexible, and hard to audit, while explicit adaptivity surfaces modular structure that is structured, auditable, and controllable. The transition highlights three key trade-offs developed in this section: Fidelity vs. Interpretability, Local vs. Global Scope, and Approximation vs. Control.

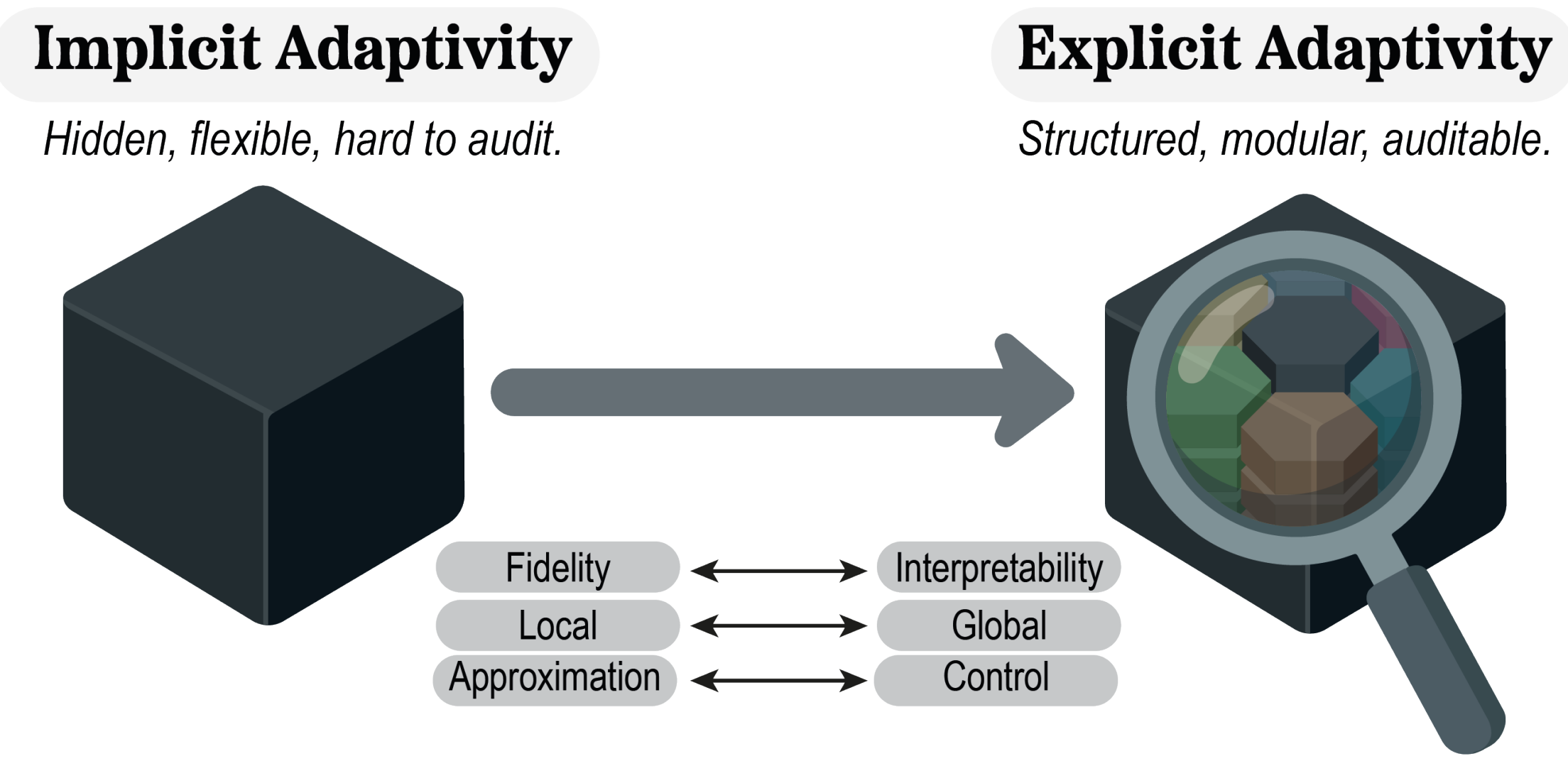

**Figure 12:** From Implicit to Explicit Adaptivity. A black-box model (left) represents implicit adaptation, which is hidden and opaque. Making adaptivity explicit (right) exposes structured components that can be inspected and controlled. The axes below highlight the trade-offs between fidelity and interpretability, local and global scope, and approximation and control.

## Approaches

Efforts to make implicit adaptation explicit span complementary strategies that differ in assumptions, granularity, and computational cost. We group them into six families:

1. surrogate modeling for local approximation,
2. prototype- and neighbor-based reasoning,
3. diagnostics for amortized inference,
4. disentanglement and bottleneck methods,
5. parameter extraction and probing, and
6. emerging approaches that leverage large language models as post-hoc explainers.

## Surrogate Modeling

This line of work approximates a black-box $h(x, c)$ with an interpretable model in a small neighborhood, so that local behavior and a local view of $f(c)$ can be inspected. A formal template is

$$\hat{g}_{x_0,c_0} = \arg\min_{g \in \mathcal{G}} \mathbb{E}_{(x,c)\sim\mathcal{N}_{x_0,c_0}} \left[\ell\big(h(x,c), g(x,c)\big)\right] + \Omega(g),$$

where $\mathcal{N}_{x_0,c_0}$ defines a locality (e.g., kernel weights), $\ell$ measures fidelity, and $\Omega$ controls complexity. Where $\mathcal{G}$ denotes a restricted hypothesis class, often composed of linear or other low-complexity functions chosen to enhance interpretability. A convenient local goodness-of-fit is

$$R^2_{\text{local}} = 1 - \frac{\sum_i w_i \left(h_i - g_i\right)^2}{\sum_i w_i \left(h_i - \bar{h}\right)^2}, \qquad w_i \propto \kappa\big((x_i, c_i), (x_0, c_0)\big),$$

where $h_i = h(x_i, c_i)$ and $g_i = g(x_i, c_i)$ are the black-box and surrogate predictions, $\bar{h}$ their weighted mean, and $\kappa$ a locality kernel centered at the point of interest $(x_0, c_0)$. LIME perturbs inputs and fits a locality-weighted linear surrogate [177]; SHAP / DeepSHAP provide additive attributions based on Shapley values [178]. Integrated Gradients and DeepLIFT link attribution to path-integrated sensitivity or reference-based contributions [179,180]. These methods are most reliable when the model is near-linear in the chosen neighborhood and perturbations remain near the data manifold; consequently, a rigorous analysis involves stating the neighborhood definition, reporting the surrogate's goodness-of-fit, and assessing stability across seeds and baselines.

## Prototype and Nearest-Neighbor Methods

Here, a decision is grounded by reference to similar cases in representation space, which supports case-based explanations and modular updates. ProtoPNet learns a library of visual prototypes to implement "this looks like that" reasoning [181]. Deep $k$-nearest neighbors audits predictions by querying neighbors in activation space and can flag distribution shift [182]. Influence functions link a prediction to influential training points for data-centric debugging [183]. This line of work connects naturally to exemplar models and contextual bandits, where decisions are justified via comparisons to context-matched exemplars. Reports include prototype coverage and diversity, neighbor quality checks, and the effect of editing prototypes or influential examples. These prototype-based approaches make local adaptation explicit by grounding predictions in reference cases, bridging the gap between black-box models and case-based reasoning frameworks.

## Amortization Diagnostics

For amortized inference systems (e.g., VAEs), the encoder $q_\phi(\theta \mid x)$ can be treated as an implicit $f(c)$. Diagnostics measure amortization gaps and identify suboptimal inference or collapse [184]. Useful outputs include calibration under shift and posterior predictive checks, together with ablations that vary encoder capacity or add limited iterative refinement. This clarifies when the learned mapping is faithful versus when it underfits the target posterior. Such diagnostics mirror classical checks for approximate Bayesian inference, where amortization gaps quantify the discrepancy between learned and exact posteriors.

## Disentangled and Bottlenecked Representations

While amortization diagnostics target model faithfulness, disentanglement aims to expose interpretable subspaces aligned with distinct contextual factors. The aim is to expose factors that align with distinct contextual causes, making changes traceable and controllable. $\beta$-VAE encourages more factorized latents [185], while the Deep Variational Information Bottleneck promotes predictive minimality that can suppress spurious context [186]. Concept-based methods such as TCAV and ACE map latent directions to human concepts and test sensitivity at the concept level [187,188]. Fully unsupervised disentanglement is often ill-posed without inductive bias or weak supervision [189]. Quantitative evaluation of disentanglement can follow established metrics that assess factor independence, completeness, and informativeness [190]. Reports should include concept validity tests, factor stability across runs, and simple interventions that demonstrate controllability.

## Parameter Extraction and Probing

This family locates where adaptation is encoded and exposes handles for inspection or edits. Linear probes test what is linearly decodable from intermediate layers [191]; edge probing examines specific linguistic structure in contextualized representations [192]. Model editing methods such as ROME can modify stored factual associations directly in weights [193], while "knowledge neurons" seek units linked to particular facts [194]. Sparse autoencoders go beyond specific neurons, extracting high-level concepts as groups of activations in language models [195]. This method also permits concept-level control of LLM behavior. Evaluation for all of these methods involves quantifying pre- and post-edit behavior, assessing locality and persistence, and documenting side effects on unrelated capabilities. Collectively, these methods transform hidden internal adaptations into analyzable modular components.

## LLMs as Post-hoc Explainers

Recent work uses in-context prompting to elicit rationales, counterfactuals, or error hypotheses from large language models for a target system [196]. These explanations can be useful but must be validated for faithfulness, for example by checking agreement with surrogate attributions, reproducing input–output behavior, and testing stability to prompt variations. Explanations should be treated as statistical estimators with stated objectives and evaluation criteria [176].

These methodological families differ in their assumptions and computational granularity, yet they all aim to render adaptation transparent and controllable. The following sections summarize their key trade-offs and conceptual challenges.

# Trade-offs

## Fidelity vs. Interpretability

High-fidelity surrogates capture the target model's behavior more accurately, yet they often grow in complexity and lose readability. A crisp statement of the design goal is

$$\min_{g \in \mathcal{G}} \underbrace{\phi_{\text{fid}}(g; U)}_{\text{faithfulness on use set } U} + \lambda \underbrace{\psi_{\text{simplicity}}(g)}_{\text{sparsity / size / semantic load}},$$

where $\phi_{\text{fid}}$ can be local $R^2$, AUC, or rank correlation with $h$, and $\psi_{\text{simplicity}}$ can be sparsity, tree depth, rule count, or active concept count. If a simple surrogate underfits, consider structured regularization (e.g., monotonic constraints, grouped sparsity, concept bottlenecks). If a complex surrogate is needed, accompany it with readable summaries (partial dependence snapshots, distilled rule sets, compact concept reports).

## Local vs. Global Scope

Local surrogates aim for $g_{x_0,c_0} \approx h$ only on $\mathcal{N}_{x_0,c_0}$, whereas a global surrogate seeks $g_{\text{global}} \approx h$ across the domain, potentially smoothing away distinct regimes. Hybrid schemes combine both:

$$g(x,c) = \sum_{k=1}^{K} w_k(x,c)\, g_k(x,c), \qquad \sum_k w_k(x,c) = 1, \quad w_k \geq 0,$$

with local experts $g_k$ and soft assignment $w_k$. Report the neighborhood definition, coverage (fraction of test cases with acceptable local fit), and disagreements between local and global views; flag regions where the global surrogate is unreliable.

### Approximation vs. Control

Coarse modularization makes control and auditing simpler because edits act on a small number of levers, yet residual error can be large. Fine-grained extraction, such as neuron- or weight-level edits, can achieve precise behavioral changes but may introduce unintended side effects. Define the intended edit surface in advance (concepts, features, prototypes, submodules, parameters). For coarse modules, measure the residual gap to the base model and verify that edits improve target behavior without harming unaffected cases. For fine-grained edits, quantify locality and collateral effects using a held-out audit suite with counterfactuals, canary tasks, and out-of-distribution probes. Maintain versioned edits, enable rollback, and document the scope of validity.

These trade-offs determine the operational boundaries where explicit representations can remain faithful to the original adaptive system.

## Evaluation and Reporting Standards for Classical Post-hoc Methods

LIME, SHAP, and gradient-based methods such as Integrated Gradients and DeepLIFT remain common tools for context-adaptive interpretation. Their usefulness depends on careful design and transparent reporting. Explanations should be treated as statistical estimators with stated objectives and evaluation criteria [175,176]. Carmichael & Scheirer (2021) further propose a principled evaluation framework for feature-additive explainers, enabling measurement of misattribution even under known ground-truth additive models [197].

### Scope and locality

Local surrogate methods require a clear definition of the neighborhood in which the explanation is valid. The sampling scheme, kernel width, and surrogate capacity determine which aspects of the black box can be recovered. When context variables are present, the explanation should be conditioned on the relevant context and the valid region should be described.

### Attribution methods in practice

Attribution based on gradients is sensitive to baseline selection, path construction, input scaling, and preprocessing. Baselines should have clear domain meaning, and sensitivity analyses should show how conclusions change under alternative baselines. For perturbation-based surrogates, report the perturbation distribution and any constraints that keep samples on the data manifold.

### Faithfulness and robustness

Faithfulness and robustness should be checked rather than assumed. Useful checks include deletion and insertion curves, counterfactual tests, randomization tests, stability under small input and seed perturbations, and for local surrogates a local goodness-of-fit such as a neighborhood $R^2$. The evaluation metric should match the stated objective of the explanation [175,176]. Turbé et al. (2023) demonstrate evaluation of interpretability methods on time-series models using metrics such as $\widetilde{\mathrm{AUC}}_S$ and $\widetilde{F_{1,S}}$ to compare alignment with model internals [198].

## Minimal reporting checklist

| Item | Description |
|---|---|
| **Data slice and context definition** | Specify the subset of data and contextual variables used for generating explanations, and describe the locality or neighborhood definition. |
| **Surrogate specification and regularization details** | Report the family of surrogate models, chosen regularization strategy, and kernel or sampling parameters. |
| **Faithfulness and robustness metrics** | Include local $R^2$, deletion/insertion area, counterfactual validity, and robustness under perturbations. |
| **Sensitivity and uncertainty analysis** | Assess variation across baselines, random seeds, and small input perturbations, providing uncertainty estimates. |
| **Computational constraints** | Document runtime, hardware limitations, and approximation budgets that affect explanation quality. |
| **Observed limitations and failure modes** | Summarize known weaknesses, unstable regions, or interpretability failures identified during validation. |

**Table 2. Minimal Reporting Checklist for Post-hoc Explanations**

## From post hoc analysis to design

Insights from post-hoc analysis can inform proactive model design for control, auditing, and policy communication. In such cases, interpretability methods should not remain external diagnostics but serve as guides for architectures with built-in transparency. For example, Concept Bottleneck Models integrate interpretable concepts into the forward pass [199]. Similarly, Poursabzi-Sangdeh et al. (2021) conduct empirical user studies to highlight how interpretability design choices affect human use and model trust [200]. These contributions extend the vision of Doshi-Velez & Kim (2017) toward a unified science of interpretable modeling, where explanation and model training are co-designed [175]. Taken together, these lines of work bridge black-box adaptation and structured inference and set the stage for designs where context-to-parameter mappings are specified, trained, and evaluated end to end.

## Implications for classical models

These tools can also clarify how traditional models, for example, logistic regression with interaction terms or generalized additive models to admit a local adaptation view: a simple global form paired with context-sensitive weights or features. Reading such models through the lens of local surrogates and concept interfaces helps align classical estimation with modern, context-adaptive practice. Reinterpreting these classical estimators through the lens of explicit adaptivity situates them as early instances of structured context modeling, underscoring continuity between statistical modeling and modern machine learning.

Taken together, these strategies illustrate a gradual unification of interpretability, modularization, and adaptive modeling, paving the way toward a principled science of explicit context-aware inference.

# 9. Context-Invariant Training: A View from the Converse

While the preceding sections emphasize the importance of modeling context to tailor predictions, an equally fundamental question concerns robustness: Can we learn representations such that a single predictor performs reliably across sites, cohorts, and time, despite environmental shifts and nuisance variation? Context-invariant training aims at out-of-distribution (OOD) generalization by emphasizing features whose associations with the target remain stable across environments, while suppressing spurious correlations that vary with nuisance contexts. Standard Empirical Risk Minimization (ERM) [201] often latches onto spurious, environment-specific correlations. In practice, this means using multiple environments during training and favoring representations that make a single readout perform well everywhere.

The seminal framework connecting modern deep learning to invariant prediction is Invariant Risk Minimization (IRM) [202], which formulates robustness as learning causally stable predictors across multiple environments. IRM seeks a representation $\Phi$ such that a shared predictor $w$ minimizes the risk $R^e(\cdot)$ for every environment $e$. The original formulation is a bi-level optimization problem that is computationally intractable. To make it solvable, Arjovsky et al. propose a surrogate version, IRMv1, which introduces a penalty ensuring that the per-environment risk gradient vanishes for a shared dummy classifier $w = 1$, thereby enforcing stationarity across environments. This construction connects invariance to out-of-distribution (OOD) generalization by encouraging predictors aligned with causal mechanisms that persist across environments.

However, subsequent analyses revealed important limitations. In linear settings, IRM often fails to recover the true invariant predictor, and in nonlinear regimes, performance can deteriorate sharply when test distributions deviate from the training domains [203]. This undermines IRM's objective of handling distribution shift, where $P(X)$ changes while $P(Y|X)$ remains fixed. Thus, IRM offers no mechanism to reduce sensitivity when those shifts are amplified at test time. To mitigate these issues, Risk Extrapolation (REx) [204] extends the principle of invariance by optimizing directly over per-environment risk vectors. Two practical variants have been proposed: MM-REx and V-REx, which performs robust optimization over affine combinations of the environment risks (weights sum to 1, possibly negative), and V-REx, which minimizes the mean risk augmented by the variance of risks across environments.

Unlike IRM, which requires explicit environment labels, Beery et al. (2018) [205] propose CoRe, a method that assumes some samples share a common identifier. Features are decomposed into core components (whose class-conditional distribution is stable across domains) and style components (e.g., brightness, pose) that vary with domains. The CoRe estimator enforces robustness by penalizing the conditional variance of the loss within groups sharing the same label–identifier pair $(Y, ID)$.

## Adversarial Robustness as Context-Invariant Training

Whereas IRM seeks robustness across discrete environments, adversarial robustness can be regarded as its infinitesimal counterpart—focusing on perturbations within a local neighborhood of each input rather than across distinct domains. Those different environments can be interpreted as fine-grained, synthetic perturbations around each data point rather than distinct real-world domains. Invariant learning generally seeks predictors whose performance remains stable when the data-generating context changes — for example, across hospitals, time periods, or demographic groups [206]. Adversarial robustness follows the same principle of invariance, but at a much finer scale: instead of using naturally occurring environments, it constructs synthetic "environments" through small, deliberate perturbations of the input data. These perturbations simulate local environmental shifts around each sample and expose the model to worst-case contexts. From this perspective, adversarial robustness is essentially context-invariant learning under infinitesimal, adversarially generated environments. Each adversarial example $x' = x + \delta$ (where $\|\delta\|_p \leq \varepsilon$) can be interpreted as belonging to a neighboring environment of the original input x. Training the model to perform consistently under such local shifts enforces a form of fine-grained invariance that complements the coarse-grained invariance targeted by IRM. The paper [207] addresses the vulnerability of deep learning models to adversarial attacks from the optimization view. Specifically, the authors interpret adversarial robustness as a min-max optimization problem, where the goal is to minimize the worst-case loss incurred by adversarial examples. Madry et al. (2018) introduce Projected Gradient Descent (PGD) as a universal first-order adversary. The generated perturbations are incorporated into the training process to improve robustness under local contextual shifts. In this view, the environments in IRM correspond to multiple data domains, while those in adversarial training correspond to local neighborhoods of each sample—both formulations share the same objective of minimizing performance variation across shifts in context. Formally, both IRM and adversarial training minimize performance variance across contexts—IRM across discrete environments, and adversarial training across continuous perturbation neighborhoods.

[208] provably demonstrates the trade-off between robustness and accuracy in machine learning models. The authors argue that adversarial training, while improving robustness to adversarial perturbations, can decrease the model's accuracy on clean data. This occurs because adversarial training forces the model to adjust its decision boundaries, which may lead to a loss in standard performance. The paper also shows that the representations learned by robust models align better with salient data characteristics and human perception, which suggests that robust models focus more on features that are meaningful and interpretable. At the same time, robust models tend to learn representations that align better with salient data characteristics and human perception, suggesting that robustness promotes the extraction of stable, semantically meaningful features, mirroring the goal of context-invariant learning at a smaller, instance-specific scale [209].

This perspective is directly applicable to the challenges faced by LLM-based Agents as surveyed in [210]. An autonomous agent does not operate in a sterile, curated dataset; it operates in the wild. These fine-grained, synthetic perturbations provide a useful abstraction for understanding the robustness challenges faced by LLM-based agents:

**Perception Robustness**: A small, imperceptible change to an image or a document an agent is analyzing (an adversarial perturbation) could cause it to completely misinterpret its environment and take a disastrous action.

**Tool-Use Robustness**: A slight rephrasing of a user's command could trick a non-robust agent into generating incorrect or malicious code for a tool to execute.

Hence, advances in adversarial robustness directly inform the design of safer, more context-stable autonomous agents.

## Training methods for Context-Invariant Models

While the principle of context-invariance is a powerful theoretical goal, several practical training methodologies have been developed to approximate it, primarily by enhancing robustness against group shifts. These methods vary in their assumptions, particularly regarding the availability of explicit group or environment labels for the training data.

A foundational approach, applicable when group labels are available, is Group Distributionally Robust Optimization (Group DRO). Unlike standard Empirical Risk Minimization (ERM) which minimizes the average loss over the entire dataset, formulated as:

$$\min_f \frac{1}{n} \sum_{i=1}^{n} L(f(x_i), y_i)$$

Group DRO's objective is to minimize the loss on the worst-performing data group. This is formally expressed as a min-max problem:

$$\min_f \max_{g \in \mathcal{G}} \mathbb{E}_{(x,y)\sim P_g}[L(f(x), y)]$$

where $\mathcal{G}$ represents the set of all predefined groups and $P_g$ is the data distribution for a specific group $g$ [211]. However, the authors identify a critical pitfall: in modern, overparameterized neural networks, this method can fail. Such models can easily memorize the entire training set, reducing the worst-case training loss to zero without actually learning a generalizable solution. The key insight from this work is that **strong regularization** (such as a high L2 penalty or aggressive early stopping) is essential. Regularization prevents the model from perfectly fitting the training data, forcing it to learn simpler, more robust features that generalize better to the worst-case groups on unseen data. The primary limitation of Group DRO is its reliance on fully annotated training data, a luxury seldom available in real-world scenarios. This challenge has spurred the development of methods that operate without explicit group information. These approaches cleverly leverage the inherent biases of standard models as a source of information. A simple and highly effective heuristic is Just Train Twice (JTT) [212]. This method operates in two stages: first, a standard ERM model is trained for several epochs. Second, the training examples that this initial model misclassified are identified and upweighted. A new model is then trained from scratch on this reweighted dataset. The underlying assumption is that a standard model's errors serve as an effective proxy for identifying examples from minority or difficult groups. By focusing the second stage of training on these hard examples, JTT improves worst-group performance without ever needing to know the group labels. Providing a more formalized framework, Environment Inference for Invariant Learning (EIIL) aims to bridge the gap between unlabeled data and invariant learning algorithms like IRM [213]. Similar to JTT, EIIL begins by training a standard ERM model. It then uses the biases of this reference model to automatically partition the dataset into several inferred "environments." For instance, examples the model confidently gets right might form one environment, while those it gets wrong form another. These algorithmically generated environment labels can then be fed into any off-the-shelf invariant learning method to train a final, robust model. EIIL essentially transforms the problem from one requiring manual labels to one where environments can be discovered directly from the data itself. Collectively, these approaches demonstrate a continuum from fully supervised environment-aware optimization to self-supervised environment discovery, unified under the goal of achieving context-invariant generalization. Together, these methods illustrate a clear progression from fully-supervised techniques to more practical approaches that cleverly infer hidden data structure, all aiming to build models that are more robust and invariant to challenging shifts in context.

# 10. Applications, Case Studies, and Tools

This section surveys how context-adaptive methods manifest across domains and what tools enable them in practice.

## Implementation Across Sectors

Many real-world environments are dynamic and unpredictable, meaning that models built on static assumptions often fail when conditions shift. To remain reliable, models must be able to adapt to changing inputs, contexts, and behaviors. This adaptability is especially important in high-stakes domains where decisions directly affect human well-being or carry significant financial consequences. Two prominent examples are healthcare and finance. In healthcare, context-adaptive models enable more personalized treatment decisions and support early intervention by capturing the evolving state of patients and diseases. In finance, these models capture rapidly changing market conditions, allowing forecasts and risk assessments to remain accurate in volatile times.

Healthcare is one of the domains that benefits greatly from context-aware models because clinical and biomedical data are often hierarchical, exhibiting nested structures and evolving over time. For example, patients may have repeated measurements (e.g., vitals, labs) nested within visits, and these visits are themselves nested within broader care episodes. At the same time, disease trajectories and treatment responses are highly dynamic, requiring models that can adapt to changing contexts rather than assuming static relationships. Several reviews highlight the importance of methods that explicitly account for such complexity in longitudinal and multilevel health data [214,215]. One concrete example is a Bayesian multilevel time-varying joint model that captures complex structures while estimating diverse time-varying relationships, including both response–predictor and response–response dependencies [216]. Such models often employ hierarchical priors to borrow strength across patients while maintaining individualized inference. In this framework, time-varying coefficients are flexibly estimated using Bayesian P-splines, and inference is performed through Markov Chain Monte Carlo (MCMC). The result is a computationally efficient algorithm that provides interpretable modeling of patient outcomes as they evolve over time.

In finance, context-aware models are particularly valuable for capturing the complex dynamics that unfold both over time and across countries, sectors, and assets, which together drive macroeconomic and market behavior. For instance, cross-sectional dependencies, which capture interconnectedness at the same point in time, emerge when shocks propagate differently across regions or industries, while temporal dependencies, which capture persistence across time, arise from persistent volatility clustering and regime changes. Several reviews and comparative studies emphasize the need for methods that can adapt to such heterogeneity in modern financial data [217,218]. A prominent line of work develops Bayesian matrix dynamic factor models (MDFMs), which provide a powerful framework for analyzing matrix-valued time series increasingly common in macro-finance applications [219]. These models incorporate multiple context-adaptive features. On the temporal side, an autoregressive factor process captures persistent comovement and improves recursive forecasting, while stochastic volatility, fat-tailed error distributions, and explicit COVID-19 outlier adjustments allow the model to remain robust under real-world market shocks. The approximate factorization reduces complexity from cubic to linear in the number of assets, making large-scale forecasting feasible.

## Context-Aware Efficiency in Practice

The principles of context-aware efficiency find practical applications across diverse domains, demonstrating how computational and statistical efficiency can be achieved through intelligent context utilization.

In healthcare applications, context-aware efficiency enables adaptive imaging protocols that adjust scan parameters based on patient context such as age, symptoms, and medical history, reducing unnecessary radiation exposure. Personalized screening schedules optimize screening frequency based on individual risk factors and previous results, while resource allocation systems efficiently distribute limited healthcare resources based on patient acuity and context.

Financial services leverage context-aware efficiency principles in risk assessment by adapting risk models based on market conditions, economic indicators, and individual borrower characteristics. Fraud detection systems use context-dependent thresholds and sampling strategies to balance detection accuracy with computational cost, while portfolio optimization dynamically adjusts rebalancing based on volatility regimes and transaction costs, as studied in regime-switching portfolio models [220].

Industrial applications benefit from context-aware efficiency through predictive maintenance systems that adapt maintenance schedules based on equipment context including age, usage patterns, and environmental conditions [221]; recent surveys of predictive maintenance in Industry 4.0 identify architectures that integrate sensor data, remaining-useful-life models, and context-aware scheduling policies [222,223]. Quality control implements context-dependent sampling strategies that focus computational resources on high-risk production batches, and inventory management uses context-aware forecasting to optimize stock levels across product categories and market conditions, outperforming traditional forecasts in retail settings [224].

A notable example of context-aware efficiency is adaptive clinical trial design, where trial parameters are dynamically adjusted based on accumulating evidence while maintaining statistical validity. Population enrichment refines patient selection criteria based on early trial results, and dose finding optimizes treatment dosages based on individual patient responses and safety profiles. These applications demonstrate how context-aware efficiency principles can lead to substantial improvements in both computational performance and real-world outcomes.

## Contextualized Network Inference

One domain where context-adaptive models have shown particular promise is in network inference for genomics. Traditional approaches assume that all samples can be pooled into a single network, or that cohorts can be partitioned into homogeneous groups. These assumptions are often unrealistic: cancer, for example, exhibits both cross-patient heterogeneity and within-patient shifts in gene regulation.

Contextualized network models address this challenge by learning archetypal networks and then representing each sample as a mixture of these archetypes, weighted by its observed context. This formulation allows researchers to move beyond average-case networks and uncover mechanisms of disease, heterogeneity across patients, driver mutations, and structural hazards.

Such contextualized networks have been applied in TCGA cancer genomics to identify patient-specific driver modules.

Context-adaptive networks unlock new views of biological regulation

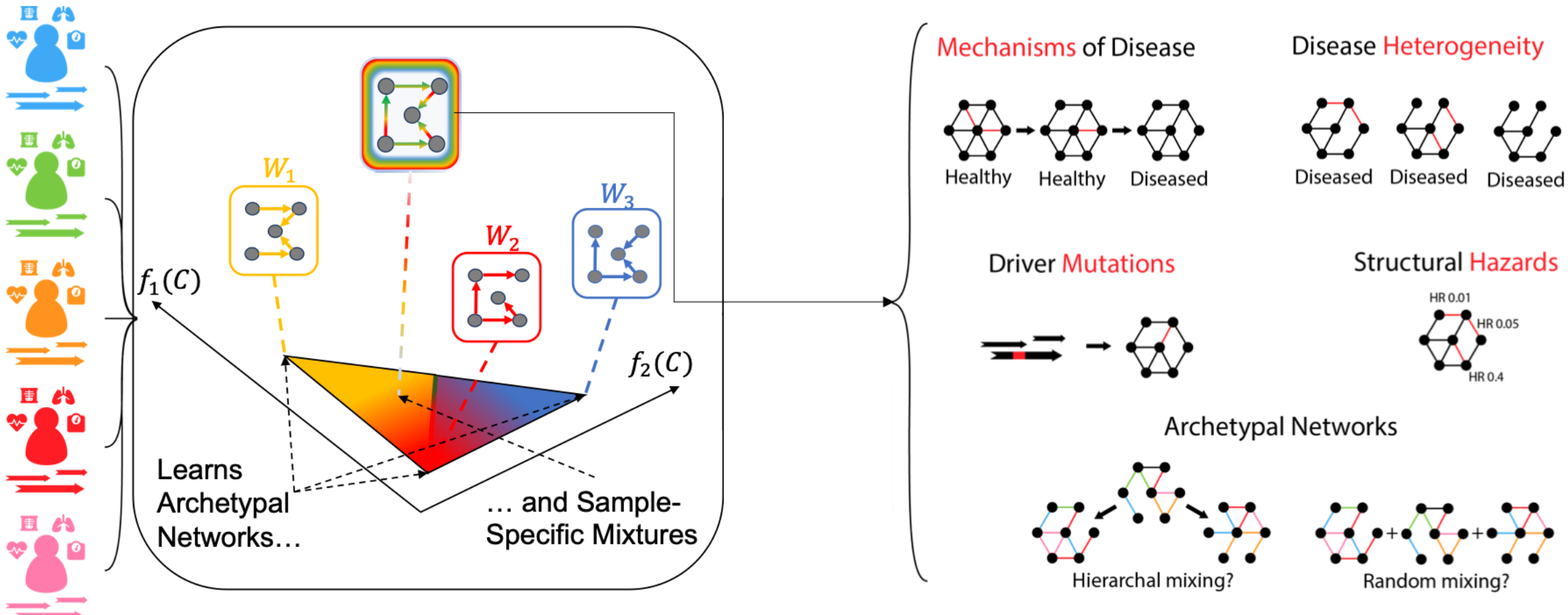

**Figure 13:** Contextualized networks enable inference of archetypal and sample-specific mixtures, unlocking new biological insights such as mechanisms of disease, disease heterogeneity, structural hazards, and driver mutations.

## Empirical Evaluation in Practice

Evaluating context-adaptive models requires careful consideration of predictive accuracy, robustness to variability, and scalability, with the emphasis varying by domain. Key aspects of performance evaluation include the choice of metrics, the handling of uncertainty, and assessment under stress or rare-event conditions.

In healthcare, evaluation prioritizes patient-specific predictive accuracy and calibrated uncertainty. Common metrics include mean squared error (MSE), concordance indices (C-index), and calibration curves, which measure how well models capture longitudinal patient trajectories and provide reliable uncertainty estimates. Multi-target Bayesian approaches and survival models demonstrate the importance of capturing correlations across outcomes and assessing credible interval coverage to quantify predictive confidence [225]. Evaluations in this domain also highlight trade-offs between model complexity, interpretability, and computational feasibility, since high-fidelity patient-level predictions can be costly to compute.

In finance and macro forecasting, performance evaluation emphasizes predictive accuracy under volatile conditions and resilience to structural breaks. Metrics such as root mean squared forecast error (RMSFE), log-likelihood, and stress-test performance are commonly used to assess how well models handle crises or abrupt shifts in data [226,227]. Probabilistic metrics, including posterior predictive checks and uncertainty bounds, provide additional insight into the reliability of forecasts, while chaos-informed diagnostics can highlight vulnerabilities to extreme events [228].

Across domains, consistent patterns emerge. Context-adaptive models outperform static baselines when variability is structured and partially predictable, but require sufficient examples of the desired adaptation behavior across contexts. Even if the dataset is large, if there are insufficient contexts to show context-specific adaptation, the model will fail to generalize across contexts. Thus, learning context-adaptive models requires special attention not only to the number of samples, but also to the number of distinct contexts and the distribution of contexts that samples are drawn from [97].

## Survey of Tools

There are many technological supports that have emerged to support context-adaptive modeling. These tools provide the infrastructure, memory, and efficiency mechanisms that allow models to operate effectively in dynamic environments.

Retrieval-augmented generation (RAG) has become a core support for adaptivity, enabling models to incorporate new knowledge at inference time instead of relying only on static parameters. Recent surveys outline how RAG architectures combine dense retrievers, re-rankers, and generators into pipelines that continuously update with external information. This allows models to remain aligned with changing knowledge bases [229]. Beyond improving factuality, RAG also underpins adaptive behavior in AI-generated content, where external retrieval reduces hallucination and provides domain-specific grounding [230]. These systems depend on efficient vector search. Tools such as FAISS use approximate nearest neighbor algorithms to index billions of embeddings with low latency, while Milvus integrates distributed storage to scale such systems across production environments [231]. Together, retrieval pipelines and vector databases constitute the infrastructure through which context-adaptive models dynamically expand their accessible knowledge.

While retrieval addresses external knowledge, memory systems support continuity within ongoing interactions. Research on AI memory frameworks emphasizes how models require mechanisms to persist relevant context, get rid of redundancy, and resurface information at appropriate times [232].

Recent implementations such as MemoryOS illustrate how adaptive memory systems can summarize past conversations, cluster related items, and strategically reinsert them into prompts, producing long-term coherence that can't be achieved with static context windows alone [233]. These memory architectures extend adaptivity from the level of just accessing facts to maintaining evolving histories, allowing models to not just adjust to new data, but also to be more consistent and contextually aware of their interactions.

Another critical support lies in scaling sequence length. Standard transformers suffer quadratic complexity and degraded performance as contexts grow, making it difficult to adapt to long or streaming data. New serving infrastructures such as StreamingLLM introduce rolling caches that let models handle long inputs without full recomputation, while frameworks like vLLM use paged attention to manage GPU memory efficiently during extended inference [234,235]. This long-context support shifts adaptability from handling snapshots of information to maintaining awareness across evolving information streams.

## Selection and Usage Guidance

Deploying context-adaptive models effectively requires careful alignment between model capabilities, domain needs, and practical constraints.

In healthcare, where data is often hierarchical and time-varying, Bayesian multilevel models and generalized varying-coefficient frameworks are well suited because they can flexibly capture nonlinear interactions and evolving patient trajectories. In finance, high-dimensional time series demand scalability, making matrix dynamic factor models more appropriate than fully specified multivariate systems.

Domain priorities should drive tool choice. Clinical applications often require interpretable models that clinicians can trust, favoring spline-based or single-index approaches even if they sacrifice some predictive accuracy. In contrast, finance applications typically prioritize forecasting performance under volatility, where more complex factor models can offer a competitive edge despite reduced transparency.

Many context-adaptive models rely on resource-intensive inference methods such as MCMC, which may limit scalability. Approximate inference techniques like variational Bayes or stochastic optimization can mitigate this burden for large datasets. In real-time decision settings, long-context processing methods such as StreamingLLM or KV-cache compression provide efficiency gains but require specialized engineering and hardware support.

Finally, tool selection should reflect whether the primary objective is scientific insight or operational decision-making. Biomedical research benefits most from flexible, interpretable models that generate new hypotheses, whereas domains like trading demand models capable of rapid adaptation, scalable inference, and strong predictive accuracy under uncertainty.

There is no one-size-fits-all context-adaptive model. Successful deployment depends not only on technical choices but also on aligning model adaptivity with domain-specific interpretability and governance requirements.

# 11. Future Trends and Opportunities with Foundation Models

## A New Paradigm for Context-Adaptive Inference

Recent advances in large-scale foundation models have fundamentally reshaped the landscape of context-adaptive inference. Trained on vast and diverse datasets with self-supervised objectives, these models internalize broad statistical regularities across language, vision, and multimodal data [60]. Unlike earlier approaches that relied on hand-crafted features or narrowly scoped models, foundation models can process and structure complex, high-dimensional contexts that were previously intractable.

Their impact is clear in natural language processing, where large language models achieve strong zero-shot and few-shot generalization, and in computer vision, where multimodal encoders such as CLIP align images and text into a shared representation space [59]. These advances mark a shift from treating feature extraction and inference as separate stages toward unified systems that function simultaneously as representation learners and adaptive engines. At the same time, challenges remain, including high computational demands, the risk of amplifying societal biases, and the difficulty of interpreting learned representations [236].

To understand their contribution to context-adaptive inference, we consider three dimensions: their role as universal context encoders, the mechanisms enabling dynamic adaptation, and their integration with formal statistical and causal reasoning.

### Universal Context Encoders

Foundation models act as general-purpose context encoders, transforming raw, unstructured data into meaningful representations without manual feature engineering. For textual data, models such as BERT learn embeddings that capture semantic and syntactic nuances, supporting tasks from classification to retrieval [58]. For visual and multimodal inputs, CLIP aligns images and text into a shared embedding space, enabling zero-shot classification and cross-modal retrieval [59].

These representations effectively serve as context variables—latent, structured features that can feed directly into statistical models. Classical approaches such as regression or causal inference can thus operate on data that would otherwise remain unstructured. This capacity forms the basis for integrating representation learning with formal frameworks of context-adaptive inference.

### Dynamic Adaptation Mechanisms

Foundation models enable dynamic adaptation primarily at inference time, allowing models to respond to new tasks without retraining. The most prominent mechanism is in-context learning (ICL), where models adapt behavior by conditioning on examples in a prompt, enabling rapid few-shot or zero-shot generalization [133].

Scaling is supported by modular architectures such as Mixture-of-Experts (MoE), which route inputs to specialized sub-networks for sparse activation, increasing capacity without proportional compute [237]. Parameter-efficient fine-tuning (PEFT) methods such as LoRA show that models can be adapted by updating less than one percent of weights, achieving near full fine-tuning performance [9]. While LoRA is traditionally learned via task-specific optimization, recent work replaces this training loop with amortized inference: given a task description, a model can directly predict the corresponding low-rank update. This shifts adaptation from an optimization problem to a learned mapping from context to parameters [135].

Together, these approaches illustrate how adaptation can be achieved both flexibly and efficiently, balancing generalization and computational constraints.

### Bridging with Statistical and Causal Reasoning

An emerging research direction integrates the representational capacity of foundation models with the rigor of statistical and causal inference. Methods such as LMPriors treat foundation models as task-specific priors, improving sample efficiency in estimation and decision making [238]. Models can also generate natural-language rationales that clarify predictions and summarize statistical findings, enhancing interpretability and transparency [196].

Consequently, foundation models serve as bridges between flexible representation learning and principled inference, offering a path toward adaptive systems that are both data-efficient and theoretically grounded.

## Next-Generation Methods for Contextualized Adaptive Inference

While current foundation models already enable impressive forms of adaptivity, the next phase of research looks toward methods that will shape the future of contextualized adaptive inference. These directions point ahead, emphasizing how models may be adapted, combined, and evaluated. The aim is not only greater power, but also more transparency and reliability in high-stakes settings. We highlight three forward-looking methodological trends: modular fine tuning and compositional adaptation, mechanistic insights into in-context learning, and new frameworks for reliability and calibration.

### Modular Fine-Tuning and Compositional Adaptation

Parameter-efficient fine-tuning approaches such as adapters and LoRA show that large models can be customized by updating only a small subset of parameters while preserving pretrained knowledge [9]. Future systems are expected to expand these ideas into compositional strategies, dynamically combining specialized modules optimized for different domains or contexts [239]. Modular reuse strategies have been proposed, including adapter libraries [146] and modular domain experts.

Recent findings suggest that merging multiple LoRA modules can even outperform full fine-tuning, signaling a paradigm where adaptation arises from modular reuse rather than retraining [240]. Compositional adaptation thus points toward building libraries of reusable context-specific skills that can be flexibly assembled for new tasks.

### In-Context Learning and Mechanistic Insights

Although in-context learning has revolutionized generalization, its internal mechanisms remain partly opaque. Evidence suggests that transformers may implement optimization-like updates during forward passes, effectively performing implicit gradient descent when processing examples [132]. Other analyses interpret ICL as implicit Bayesian inference, where the prompt provides evidence that reshapes the predictive distribution [241].

Mechanistic studies further identify induction heads within transformer attention circuits as critical components for pattern induction and few-shot generalization [125]. Such insights are expected to inspire architectures that enhance both transparency and stability in adaptive learning.

### Reliability, Calibration, and Context-Sensitive Evaluation

As adaptive models become more flexible, ensuring calibration and reliability across shifting contexts becomes crucial. Deep neural networks, including LLMs, are often miscalibrated, producing overconfident probabilities misaligned with true accuracy [242].

Future research will increasingly embed uncertainty quantification into adaptive pipelines through deep ensembles, Bayesian ensembling, or conformal prediction to produce valid confidence intervals [243,244]. Evaluation protocols must also stress robustness under distributional shifts, testing whether models can sustain performance and express uncertainty under novel or adversarial conditions [142,245].

By embedding calibration and robustness within design, adaptive inference can evolve toward a more trustworthy, auditable, and context-aware standard.

## Expanding Frameworks with Foundation Models

Foundation models refer to large-scale, general-purpose neural networks, predominantly transformer-based architectures, trained on vast datasets using self-supervised learning [60]. Their flexibility, scalability, and cross-domain generalization have transformed statistical modeling and data analysis.

LLMs such as GPT-4 [246] and LLaMA-3.1 [247] exemplify this progress, achieving state-of-the-art results in language understanding, summarization, and reasoning. Beyond NLP, foundation models extend to multimodal tasks [59], text embeddings [58], and even tabular and structured data [248].

Adaptivity in these systems is largely realized through prompting, which conditions responses on user-provided context without additional fine-tuning [249]. Meanwhile, Mixture-of-Experts (MoE) architectures enhance scalability by routing computation to relevant submodels for efficiency [237].

### Foundation Models as Context

Foundation models offer significant opportunities by supplying context-aware information that enhances various stages of statistical modeling and inference:

**Feature Extraction and Interpretation:** Foundation models transform raw, unstructured data into structured and interpretable representations. For example, targeted prompts enable LLMs to extract insightful features from text, providing meaningful insights and facilitating interpretability [252]. This allows statistical models to operate directly on semantically meaningful features rather than on raw, less interpretable data.

**Contextualized Representations for Downstream Modeling:** Foundation models produce adaptable embeddings and intermediate representations useful as inputs for downstream models, such as decision trees or linear models [133]. These embeddings significantly enhance the training of both complex, black-box models [253] and simpler statistical methods like n-gram-based analyses [254], thereby broadening the application scope and effectiveness of statistical approaches.

**Post-hoc Interpretability:** Foundation models support interpretability by generating natural-language explanations for decisions made by complex models. This capability enhances transparency and trust in statistical inference, providing clear insights into how and why certain predictions or decisions are made [255].

## Recent Innovations and Outlook

Several new architectures exemplify how foundation models advance context-sensitive inference through modularity and interpretability:

**FLAN-MoE** (Fine-tuned Language Model with Mixture of Experts) [256] combines instruction tuning with expert selection, dynamically activating relevant sub-models based on the context. This method significantly improves performance across diverse NLP tasks, offering superior few-shot and zero-shot capabilities. It also facilitates interpretability through explicit expert activations. Future directions may explore advanced expert-selection techniques and multilingual capabilities.

**LMPriors** (Pre-Trained Language Models as Task-Specific Priors) [238] leverages semantic insights from pre-trained models like GPT-3 to guide tasks such as causal inference, feature selection, and reinforcement learning. This method markedly enhances decision accuracy and efficiency without requiring extensive supervised datasets. However, it necessitates careful prompt engineering to mitigate biases and ethical concerns.

**Mixture of In-Context Experts** (MoICE) [257] introduces a dynamic routing mechanism within attention heads, utilizing multiple Rotary Position Embeddings (RoPE) angles to effectively capture token positions in sequences. MoICE significantly enhances performance on long-context sequences and retrieval-augmented generation tasks by ensuring complete contextual coverage. Efficiency is achieved through selective router training, and interpretability is improved by explicitly visualizing attention distributions, providing detailed insights into the model's reasoning process.

Collectively, these directions suggest a future in which foundation models evolve from passive representation learners into active, context-sensitive inference engines that unify adaptivity, efficiency, and interpretability within a principled framework.

# 12. Open Problems

Rapid advances in context-adaptive modeling have created unprecedented opportunities while revealing fundamental challenges. This chapter identifies the central methodological questions and the broader ethical and societal challenges that will shape the future trajectory of context-adaptive inference. We begin by examining four interrelated technical questions—on modularity, the benefits of explicit structure, theoretical and practical barriers, and interpretability trade-offs—that together define the frontier of adaptive modeling research. We then turn to the broader outlook, focusing on the ethical and societal implications of deploying these powerful adaptive systems.

## Open Research Questions

Recent advances have broadened the scope of adaptive inference, but many questions remain unresolved. These open problems span four domains: (i) modularity and reusability of adaptive components, (ii) the conditions under which explicit structure improves robustness and generalization, (iii) theoretical, computational, and data-related barriers to adoption, and (iv) the tension between interpretable-by-design and post-hoc interpretability. Together, these questions delineate a research agenda that bridges theoretical statistics, machine learning, and applied modeling, combining methodological depth with practical impact.

First, researchers need to examine whether skills and routines can be modularized in a way that allows portability across tasks without interference. Second, the field must clarify under what conditions explicit structure provides measurable benefits. Third, adoption is limited by both theoretical and practical barriers, including identifiability, generalization, and computational feasibility. Finally, the community must address the tension between building models that are interpretable from the start and those that rely on post-hoc explanations. The following subsections provide a more detailed discussion of these four questions.

### Can Reusable Modules Enable Portability Across Tasks?

A central question is whether the skills or routines acquired by large models can be isolated and reused as portable modules across tasks without reducing overall performance [60]. The vision of modularity is to build an ecosystem of specialized components that can be composed when needed, instead of training a new large model for each task. Concretely, a reusable module should satisfy portability, isolation, composability, and stability. Promising approaches operate at different levels: (i) representation-level constraints such as concept bottlenecks enforcing human-understandable features; (ii) memory-based mechanisms such as prototype libraries for case retrieval; and (iii) architecture-level designs such as sparse adapters or routing networks that activate context-relevant modules [147].

Applications illustrate the promise of this research. In healthcare, diagnostic modules could be reused across diseases. In natural language processing, syntax-aware modules might be applied across languages. However, modularity also introduces risks: interactions between modules may cause interference or instability in generalization, and poorly aligned components may propagate or amplify existing biases. Future work should therefore design evaluation protocols that test not only portability and composability, but also isolation of unintended side effects and robustness to distribution shift [247].

## What Are the Theoretical and Practical Benefits of Explicit Structure?

Clarifying the theoretical and practical benefits of explicit structure is an important open question. Implicit adaptation is highly flexible, but explicit structure may provide stronger guarantees of robustness and generalization under distribution shift. Benefits are most likely when domain priors are reliable, data are scarce, or safety constraints limit free-form behavior. Explicit structure is promising when context topology is known (spatial or graph), when spurious correlations should be suppressed, and when explanations must be auditable. Practical benefits include greater interpretability, improved debugging, and the ability to incorporate domain knowledge directly.

To advance this agenda, systematic comparisons with implicit approaches are needed. Stress testing under covariate shift, concept drift, long-tail distributions, and adversarial correlations is particularly important, and benchmarks such as WILDS provide a useful starting point [142]. At the same time, researchers must weigh the costs of explicit structure. These costs include additional annotation, increased hyperparameter complexity, and potential reductions in in-domain accuracy [157]. A comprehensive evaluation framework that quantifies both theoretical guarantees and practical trade-offs remains to be established.

## What Theoretical and Practical Barriers Remain?

Several barriers continue to limit the adoption of adaptive models. On the theoretical side, researchers have yet to establish strong guarantees for identifiability and generalization under distribution shift [157]. Extending these guarantees to high-dimensional and multimodal data remains an unsolved challenge.

Practical barriers are equally important. Training and deploying adaptive models requires significant computational and memory resources. Data limitations, such as biased sampling and noisy feedback, reduce reliability. Evaluation frameworks remain centered on accuracy, with insufficient attention to fairness, stability, and long-term robustness. Finally, the absence of standardized tools and implementation guidelines prevents many practitioners from applying state-of-the-art methods beyond research settings [63].

## Interpretable-by-Design vs Post-hoc Interpretability: What Is the Right Path Forward?

A final open question concerns the balance between interpretable-by-design approaches and post-hoc interpretability. Interpretable-by-design models, such as varying coefficient models, provide transparency and faithfulness from the outset but may restrict predictive performance [3]. Post-hoc methods allow powerful foundation models to be explained after training, but explanations may be incomplete or unfaithful to the model's internal reasoning [60,195].

Progress in both directions suggests that the future lies in integration rather than a binary choice. Hybrid models may embed interpretable structures at their core while using post-hoc tools for flexibility. Promising directions include benchmarks that jointly evaluate adaptivity and interpretability, as well as human-in-the-loop workflows that allow domain experts to constrain and validate model adaptation in practice.

## Broader Challenges and Future Outlook

Emerging paradigms such as Agentic Context Engineering (ACE) push this vision further by treating the context itself as an adaptive, evolving entity. In this framework, language models continuously refine and regenerate their own contexts through feedback, reflection, and planning, enabling self-improving adaptation cycles across time [258]. While the previous section focused on research questions that can be addressed by new methods, theory, and experiments, broader challenges remain that extend beyond purely technical considerations. These challenges concern the responsible deployment of adaptive models in real-world environments, where issues such as ethics, fairness, and regulatory compliance play a critical role. Adaptive systems used in sensitive domains such as healthcare and finance must satisfy principles of interpretability, auditability, and accountability to prevent harm and maintain public trust [60]. Collaboration between regulators, practitioners, and researchers is essential to establish transparent auditing standards and verifiable documentation for adaptive decisions.

Another set of challenges arises from the dynamic interaction between adaptive models and their environments. Feedback loops may amplify small initial biases, leading to systematic disadvantages for certain groups over time. Examples can be seen in credit scoring, hiring, and online recommendation systems, where early decisions influence future data collection and can entrench inequalities [159]. Addressing these risks requires methods that anticipate long-term effects, including simulation studies, formal analyses of dynamic systems, and model designs that incorporate fairness constraints directly during learning.

Looking ahead, the long-term vision for adaptive modeling is to develop systems that are not only powerful but also trustworthy. Progress requires moving beyond accuracy as the dominant evaluation criterion to include fairness, stability, and transparency. Human oversight should be an integral part of adaptive pipelines, enabling experts to guide and validate model behavior in practice. Sustainability is another important dimension, as the computational and environmental costs of adaptive models continue to grow. By combining technical innovation with responsible deployment, the field can ensure that adaptive inference contributes to both scientific progress and societal benefit.

# 13. Conclusion

## Overview of Insights

This review established a unifying framework for understanding context-adaptive inference across both explicit statistical models and implicit adaptation in modern foundation models. By tracing how adaptation appears in parameterized functions such as varying-coefficient models and in emergent processes like in-context learning, we showed that these paradigms share a common estimator form and theoretical foundation.

Across the literature, a consistent pattern emerges: adaptivity becomes effective when context, computation, and interpretation are aligned. The principles of context-aware efficiency integrate these aspects, clarifying when adaptation enhances robustness and when it introduces instability. Within this perspective, model design choices can be connected to measurable outcomes such as data efficiency, modularity, and transferability, grounding the abstract notion of adaptivity in verifiable performance.

The unified view presented in this review connects statistical inference with ideas from machine learning and cognitive modeling, where adaptive reasoning and context-sensitive generalization are regarded as key components of intelligent behavior. Cognitive theories have long emphasized that efficient adaptation arises from internal models that balance precision and flexibility, an idea now mirrored in recent computational analyses of in-context learning [126]. By bridging these perspectives, this framework provides both a conceptual foundation and a practical guide for developing adaptive systems that are interpretable, reliable, and scalable.

## Context-Aware Efficiency: A Unifying Framework

The principles of context-aware efficiency emerge as a unifying theme across the diverse methods surveyed in this review. This framework provides a systematic approach to designing methods that are both computationally tractable and statistically principled.

Several fundamental insights emerge from our analysis. Rather than being a nuisance parameter, context provides information that can be leveraged to improve both statistical and computational efficiency. Methods that adapt their computational strategy based on context often achieve better performance than those that use fixed approaches. The design of context-aware methods requires careful consideration of how to balance computational efficiency with interpretability and regulatory compliance.

Recent studies also demonstrate that context-adaptive strategies can emerge spontaneously in large models trained on diverse tasks, linking computational efficiency to rational inference principles [124]. These findings suggest that implicit adaptation can serve as a computational analog of Bayesian updating, where context dynamically reweights prior knowledge to improve generalization. Similar ideas have been explored in meta-learning frameworks such as MetaICL, which meta-trains language models to acquire reusable adaptation strategies through exposure to varied task distributions [117].

Future research in context-aware efficiency should focus on developing methods that can efficiently handle high-dimensional, multimodal context information, creating systems that can adaptively allocate computational resources based on context complexity and urgency, investigating how efficiency principles learned in one domain can be transferred to others, and ensuring that context-aware efficiency methods can be deployed in regulated environments while maintaining interpretability [258].

The development of context-aware efficiency principles has implications beyond statistical modeling. More efficient methods reduce computational costs and environmental impact, enabling sustainable computing practices. Efficient methods also democratize AI by enabling deployment of sophisticated models on resource-constrained devices. Furthermore, context-aware efficiency enables deployment of personalized models in time-critical applications, supporting real-time decision making.

As we move toward an era of increasingly personalized and context-aware statistical inference, the principles outlined in this review provide a foundation for developing methods that are both theoretically sound and practically useful.

# Future Directions

Looking ahead, the evolution of context-adaptive inference will likely proceed along four interconnected paths.

## Theoretical Foundations

Future research should formalize implicit adaptation within a consistent statistical framework, linking neural computation to principles of efficiency, identifiability, and invariance. Clarifying these theoretical connections will support better understanding of when implicit adaptation approximates explicit statistical reasoning and how both approaches can be integrated. Recent advances have begun to view in-context learning as an emergent form of structure induction, suggesting that large models implicitly learn compositional representations that approximate rational inference processes [124].

## Modular and Compositional Methods

Progress in parameter-efficient fine-tuning, compositional adaptation, and reusable modules will make large models more flexible and controllable. Building libraries of specialized components that can be dynamically combined will promote efficient reuse and domain transfer while maintaining interpretability. Work on tabular in-context learning, such as the TabICL architecture, illustrates how these principles can scale to structured data domains while preserving modular control and generalization [118].

## Evaluation and Reliability

Developing standardized benchmarks that jointly assess robustness, calibration, and interpretability is essential for advancing both theory and application. Future evaluation frameworks should emphasize context-stratified performance, long-term stability, and transparent reporting of adaptation behavior under distribution shifts. Ongoing analyses of the stability and transience of in-context strategies [126] underscore the importance of evaluating not only short-term generalization but also the persistence and reproducibility of adaptive behavior across training regimes.

## Responsible and Sustainable Deployment

As adaptive systems become embedded in decision-making processes, integrating fairness auditing, human oversight, and energy efficiency into their design will be critical for ensuring public trust. Addressing the environmental cost of large-scale adaptation and developing resource-conscious algorithms will also contribute to sustainable computing practices. Emerging work on efficient foundation models and rational adaptation frameworks [258] highlights how technical design and ethical responsibility can be jointly optimized in real-world deployment.

Together, these directions outline a path toward the next generation of adaptive models that are both powerful and trustworthy. Progress will depend on combining rigorous statistical understanding with transparent design and responsible deployment, moving steadily toward the broader goal of making implicit adaptation explicit and accountable.

# Appendix A

This appendix gives full proofs for Proposition 1 and Corollary 1. We keep the weighted support-set notation from the Introduction and make all linear-algebra steps explicit.

## A.0 Preliminaries and identities

- **Joint features.** For any pair $(x, c)$, define
$$\psi(x,c) := x \otimes \phi(c) \in \mathbb{R}^{d_x d_c}.$$
For each indexed training example $a$ (standing in for $(i, j)$), write $\psi_a := \psi(x_a, c_a)$.

- **Design/labels/weights.** Stack $N = \sum_i m_i$ training rows:
$$Z \in \mathbb{R}^{N \times d_x d_c} \text{ with rows } Z_a = \psi_a^T, \qquad y \in \mathbb{R}^N, \qquad W = \operatorname{diag}(w_a) \in \mathbb{R}^{N \times N},\ w_a \geq 0.$$
Define the (unweighted) Gram matrix $K := ZZ^\top$ and the weighted Gram
$$K_W := W^{1/2} K\, W^{1/2} \ = \ W^{1/2} Z Z^\top W^{1/2}.$$
For a query $(x, c)$, let $k(\cdot, (x, c)) := Z\,\psi(x, c) \in \mathbb{R}^N$ and $k_{(x,c)} := W^{1/2} k(\cdot, (x, c))$.

- **Vectorization identity.** For conformable matrices $A, B, C$,
$$\operatorname{vec}(ABC) = \big(C^\top \otimes A\big)\operatorname{vec}(B), \quad \langle \operatorname{vec}(B),\, x \otimes z\rangle = x^\top B z.$$

- **Weighted ridge solution.** For any $X \in \mathbb{R}^{N \times p}$, ridge objective
$$\min_\beta \ \|W^{1/2}(y - X\beta)\|_2^2 + \lambda \|\beta\|_2^2$$
has unique minimizer $\widehat{\beta} = (X^\top W X + \lambda I)^{-1} X^\top W y$ and equivalent dual form
$$\widehat{\beta} = X^\top W^{1/2} \big(W^{1/2} X X^\top W^{1/2} + \lambda I\big)^{-1} W^{1/2} y.$$
Predictions for a new feature vector $x_\star$ equal
$$\widehat{f}(x_\star) = x_\star^\top \widehat{\beta} \ = \ \underbrace{\big(W^{1/2} X x_\star\big)^\top}_{k_\star^\top} \big(W^{1/2} X X^\top W^{1/2} + \lambda I\big)^{-1} W^{1/2} y.$$
This is **kernel ridge regression** (KRR) with kernel $K_W = W^{1/2} X X^\top W^{1/2}$ and query vector $k_\star = W^{1/2} X x_\star$.

## A.1 Proof of Proposition 1(A): explicit varying-coefficients ⇔ weighted KRR on joint features

Assume the linear, squared-loss setting with $y = \langle \theta(c), x\rangle + \varepsilon$ and $\mathbb{E}[\varepsilon] = 0$.
Let the varying-coefficients model be $\theta(c) = B\,\phi(c)$ with $B \in \mathbb{R}^{d_x \times d_c}$ and ridge penalty $\lambda \|B\|_F^2$.

**Step 1 (reduce to ridge in joint-feature space).**
Vectorize $B$ as $\beta = \operatorname{vec}(B) \in \mathbb{R}^{d_x d_c}$.
By the identity above,

$$x_a^T B\,\phi(c_a) = \langle \beta,\, x_a \otimes \phi(c_a)\rangle = \langle \beta,\, \psi_a\rangle.$$

Thus the weighted objective specialized from Eq. 1 is

$$\min_{\beta \in \mathbb{R}^{d_x d_c}} \left\| W^{1/2}\big(y - Z\beta\big)\right\|_2^2 + \lambda\|\beta\|_2^2,$$

which is exactly weighted ridge with design $X \equiv Z$.

**Step 2 (closed form and prediction).**
By the ridge solution,

$$\widehat{\beta} = (Z^T W Z + \lambda I)^{-1} Z^T W y,$$

and the prediction at a query $(x, c)$ with joint feature $\psi(x, c)$ is

$$\widehat{y}(x,c) = \psi(x,c)^T \widehat{\beta} = \underbrace{\big(W^{1/2} Z\,\psi(x,c)\big)}_{k_{(x,c)}}^{\top} \big(W^{1/2} Z Z^{\top} W^{1/2} + \lambda I\big)^{-1} W^{1/2} y.$$

**Step 3 (kernel form).**
Since $K := ZZ^T$ and $K_W := W^{1/2} K W^{1/2}$,

$$\boxed{\widehat{y}(x,c) \;=\; k_{(x,c)}^T \big(K_W + \lambda I\big)^{-1} W^{1/2} y}.$$

Moreover, the $(a, b)$-th entry of the kernel matrix $K$ is

$$K_{ab} = \langle \psi_a, \psi_b\rangle = \big\langle x_a \otimes \phi(c_a),\, x_b \otimes \phi(c_b)\big\rangle = \langle x_a, x_b\rangle \cdot \langle \phi(c_a), \phi(c_b)\rangle,$$

so (A) is precisely **KRR on joint features** with sample weights $W$.
This proves part (A). ■

---

## A.2 Proof of Proposition 1(B): linear ICL ⇒ kernel regression

We analyze a single attention layer operating on the weighted support set $S(c)$, using **linear** maps for queries, keys, and values:

$$q(x,c) = Q\,\psi(x,c), \qquad k_a = K\,\psi_a, \qquad v_a = V\,\psi_a,$$

with $Q \in \mathbb{R}^{d_q \times d_\psi}$, $K \in \mathbb{R}^{d_k \times d_\psi}$, $V \in \mathbb{R}^{d_v \times d_\psi}$, $d_\psi = d_x d_c$. Let the **unnormalized** attention score for index $a$ be

$$s_a(x,c) := w_a\,\langle q(x,c), k_a\rangle \;=\; w_a\,\psi(x,c)^T Q^T K\,\psi_a.$$

Define normalized weights $\alpha_a(x,c) := s_a(x,c)/\sum_b s_b(x,c)$ (or any fixed positive normalization; the form below is pointwise in $\{\alpha_a\}$). The context representation and scalar prediction are

$$z(x,c) = \sum_a \alpha_a(x,c)\,v_a, \qquad \widehat{y}(x,c) = u^T z(x,c).$$

We prove two statements: **(B1)** exact KRR if the attention maps are fixed and only the readout is trained, and **(B2)** kernel regression with the NTK if the attention parameters are trained in the linearized regime.

### A.2.1 (B1) Fixed attention, trained linear head ⇒ exact KRR

Assume $Q, K, V$ are fixed functions (pretrained or chosen a priori), hence $\alpha_a(x, c)$ are **deterministic** functions of $(x, c)$ and the support set. Define the induced **feature map**

$$\varphi(x, c) := \sum_a \alpha_a(x, c)\, v_a \ \in \ \mathbb{R}^{d_v}.$$

Stack $\varphi_a := \varphi(x_a, c_a)$ row-wise into $\Phi \in \mathbb{R}^{N \times d_v}$. Training only the readout $u$ with weighted ridge,

$$\widehat{u} \in \arg\min_u \ \|W^{1/2}(y - \Phi u)\|_2^2 + \lambda \|u\|_2^2$$

yields $\widehat{u} = (\Phi^T W \Phi + \lambda I)^{-1} \Phi^T W y$ and predictions

$$\widehat{y}(x, c) = \varphi(x, c)^T \widehat{u} = \underbrace{\big(W^{1/2} \Phi\, \varphi(x, c)\big)}_{k_{(x,c)}}^{T} \big(W^{1/2} \Phi \Phi^T W^{1/2} + \lambda I\big)^{-1} W^{1/2} y.$$

Therefore,

$$\boxed{\widehat{y}(x, c) = k_{(x,c)}^T \big(K_W + \lambda I\big)^{-1} W^{1/2} y}, \quad K_W := W^{1/2} \underbrace{(\Phi \Phi^T)}_{=:K} W^{1/2},$$

which is exactly **kernel ridge regression** with kernel

$$k\big((x, c), (x', c')\big) = \langle \varphi(x, c), \varphi(x', c') \rangle.$$

Because $v_a = V \psi_a$ and $\alpha_a(x, c) \propto w_a\, \psi(x, c)^T Q^T K \psi_a$, $\varphi$ is a linear transform of a **weighted average of joint features**; hence the kernel is a dot-product on linear transforms of $\{\psi_a\}$. This proves (B1). ■

### A.2.2 (B2) Training attention in the linearized/NTK regime ⇒ kernel regression with NTK

Now let $\theta = (Q, K, V, u)$ be trainable, and suppose training uses squared loss with gradient flow (or sufficiently small steps) starting from initialization $\theta_0$. The **linearized model** around $\theta_0$ is the first-order Taylor expansion

$$\widehat{y}_\theta(x, c) \ \approx \ \widehat{y}_{\theta_0}(x, c) + \nabla_\theta \widehat{y}_{\theta_0}(x, c)^T (\theta - \theta_0) =: \widehat{y}_{\theta_0}(x, c) + \phi_{\text{NTK}}(x, c)^T (\theta - \theta_0),$$

where $\phi_{\text{NTK}}(x, c) := \nabla_\theta \widehat{y}_{\theta_0}(x, c)$ are the **tangent features**. Standard NTK results (for squared loss, gradient flow, and linearization-validity conditions) imply that the learned function equals **kernel regression with the NTK**:

$$k_{\text{NTK}}\big((x, c), (x', c')\big) := \big\langle \phi_{\text{NTK}}(x, c),\ \phi_{\text{NTK}}(x', c') \big\rangle,$$

i.e., predictions have the KRR form with kernel $K_{\text{NTK}}$ on the training set (and explicit ridge if used, or implicit regularization via early stopping).

It remains to identify the structure of $\phi_{\text{NTK}}$ for our **linear attention** block and show it lies in the span of **linear transforms of joint features**. Differentiating $\widehat{y}(x, c) = u^T \sum_a \alpha_a(x, c)\, V \psi_a$ at $\theta_0$ yields four groups of terms:

- **Readout path ($u$).** $\partial \widehat{y} / \partial u = \sum_a \alpha_a(x, c)\, V \psi_a = \varphi_0(x, c)$. This is linear in $\{\psi_a\}$.

- **Value path ($V$).** $\partial \widehat{y} / \partial V = \sum_a \alpha_a(x, c)\, u\, \psi_a^T$. This contributes terms of the form $(u \otimes I) \sum_a \alpha_a(x, c) \psi_a$, i.e., linear in $\{\psi_a\}$.

- **Query/key paths ($Q, K$).** For linear attention with scores $s_a = w_a\, \psi(x,c)^T Q^T K \psi_a$ and normalized $\alpha_a = s_a / \sum_b s_b$, derivatives of $\alpha_a$ w.r.t. $Q$ and $K$ are linear combinations of $\psi(x,c)$ and $\{\psi_a\}$:

  $$\frac{\partial \alpha_a}{\partial Q} \propto \sum_b \left[\delta_{ab} - \alpha_b(x,c)\right] w_a w_b \big(K\psi_a\, \psi(x,c)^T\big), \qquad \frac{\partial \alpha_a}{\partial K} \propto \sum_b \left[\delta_{ab} - \alpha_b(x,c)\right] w_a w_b \big(\psi(x,c)\, \psi_a^T Q$$

  and hence $\partial \widehat{y}/\partial Q$, $\partial \widehat{y}/\partial K$ are finite linear combinations of tensors each bilinear in $\psi(x,c)$ and some $\psi_a$. Contracting with $u$ and $V$ produces terms linear in $\psi(x,c)$ and linear in the set $\{\psi_a\}$.

Collecting all components, the tangent feature map can be written as

$$\phi_{\mathrm{NTK}}(x,c) = \mathcal{L}\big(\psi(x,c), \{\psi_a\}\big),$$

where $\mathcal{L}$ is a fixed linear operator determined by $\theta_0$, $W$, and the normalization rule for attention. Consequently, the NTK takes the **dot-product** form

$$k_{\mathrm{NTK}}\big((x,c),(x',c')\big) = \Psi(x,c)^T\, \mathcal{M}\, \Psi(x',c'),$$

for some positive semidefinite matrix $\mathcal{M}$ and a finite-dimensional feature stack $\Psi$ that concatenates linear transforms of $\psi(x,c)$ and of the support-set $\{\psi_a\}$. In particular, $k_{\mathrm{NTK}}$ is a dot-product kernel on **linear transforms of the joint features** (possibly augmented by normalization-dependent combinations). Therefore, training the linear-attention ICL model in the linearized regime equals kernel regression with such a kernel—completing (B2). ■

**Assumptions for A.2.2.** Squared loss; gradient flow (or sufficiently small steps); initialization independent of the data; and a regime where the linearization error stays controlled over training (e.g., small learning rate, sufficient width/depth so that the NTK remains close to its initialization).

---

## A.3 Proof of Corollary 1: retrieval/gating/weighting as kernel/measure choices

In both A.1 and A.2, predictions have the KRR form

$$\widehat{y}(x,c) = k_{(x,c)}^T \big(K^\sharp + \lambda I\big)^{-1} \mu,$$

where $K^\sharp$ is a positive semidefinite kernel matrix computed over the support set (e.g., $K_W = W^{1/2} Z Z^T W^{1/2}$ in A.1 or $W^{1/2}\Phi\Phi^T W^{1/2}$ / $K_{\mathrm{NTK}}$ in A.2), $k_{(x,c)}$ is the associated query vector, and $\mu = W^{1/2} y$ (or an equivalent reweighting).

- **Retrieval $R(c)$ / gating.** Changing the support set $S(c)$ (e.g., via a retriever or a gating policy) **removes or adds rows/columns** in $K^\sharp$ and entries in $k_{(x,c)}$. This is equivalent to changing the **empirical measure** over which the kernel smoother is computed (i.e., which samples contribute and how).

- **Weights $w_{ij}(c)$.** Changing the weights modifies $W$ and hence replaces $K$ by $K_W = W^{1/2} K W^{1/2}$ and $k$ by $k_{(x,c)} = W^{1/2} k$. This is standard **importance weighting** in kernel regression.

- **Induced kernels.** Attention, value projections, or learned encoders change the **feature map** (e.g., $\psi \mapsto V\psi$ or $\psi \mapsto Q\psi$), thereby changing the kernel $k((x,c),(x',c')) = \langle \Phi(x,c), \Phi(x',c') \rangle$.

Thus retrieval/gating instantiate **neighborhood selection** (measure choice), and value/query/key processing instantiate **kernel choice**. ■

## A.4 Remarks

- **No Gaussianity is required.** Part (A) only uses squared loss and linear algebra; the noise model $y = f(x, c) + \varepsilon$ with $\mathbb{E}[\varepsilon] = 0$ suffices.
- **Early stopping vs. explicit ridge.** If training uses early stopping rather than explicit $\lambda$, the resulting predictor is still a kernel regressor with an implicit regularization parameter controlled by stopping time (for gradient flow on squared loss).
- **Multiple layers / nonlinear value stacks.** With deeper nonlinear stacks, the exact identities above become local/first-order (linearized) approximations; the NTK statement continues to apply under its usual conditions.